\documentclass[a4paper]{article}
\usepackage{graphicx}
\usepackage{amsmath}
\usepackage{hyperref}
\usepackage{csquotes}
\usepackage{fullpage} 
\usepackage{comment}
\usepackage{booktabs}
\usepackage{subcaption}
\usepackage{hyperref}
\usepackage{booktabs}
\usepackage{siunitx}
\usepackage{multirow}
\usepackage{makecell}
\usepackage{xcolor} 
\usepackage{authblk}
\title{Trustworthy AI for Medicine: Continuous Hallucination Detection and Elimination with CHECK}
\author[1]{Carlos Garcia-Fernandez}
\author[1]{Luis Felipe}
\author[1]{Monique Shotande}
\author[1]{Muntasir Zitu}
\author[1]{Aakash Tripathi}
\author[1]{Ghulam Rasool}
\author[1]{Issam El Naqa}
\author[2]{Vivek Rudrapatna}
\author[1]{Gilmer Valdes}

\affil[1]{Department of Machine Learning, Moffitt Cancer Center, Tampa, Florida, USA.} 
\affil[2]{Department of Medicine, University of California San Francisco, San Francisco, California, USA.} 

\begin{document}

\maketitle

\begin{quote}
``If you tell the truth you don't have to remember anything.''\\
--- Mark Twain \\
\end{quote}

\textbf{Abstract}

Large language models (LLMs) show promise in healthcare, but hallucinations remain a major barrier to clinical use. We present CHECK, a continuous-learning framework that integrates structured clinical databases with a classifier grounded in information theory to detect both factual and reasoning-based hallucinations. Evaluated on 1,500 questions from 100 pivotal clinical trials, CHECK reduced LLama3.3-70B-Instruct hallucination rates from 31\% to 0.3\% - making an open source model state of the art. Its classifier generalized across medical benchmarks, achieving AUCs of 0.95–0.96, including on the MedQA (USMLE) benchmark and HealthBench realistic multi-turn medical questioning. By leveraging hallucination probabilities to guide GPT-4o's refinement and judiciously escalate compute, CHECK boosted its USMLE passing rate by 5 percentage points, achieving a state-of-the-art 92.1\%. By suppressing hallucinations below accepted clinical error thresholds, CHECK offers a scalable foundation for safe LLM deployment in medicine and other high-stakes domains.

\section{Introduction}

The integration of large language models (LLMs) into clinical practice has opened transformative opportunities for patient care, therapeutic decision-making, and biomedical research \cite{Esteva2021}. These cutting-edge systems can process and synthesize extensive medical literature, offering real-time insights into disease pathophysiology, diagnostic pathways, and treatment options \cite{Biswas2023}. However, despite this promise, a critical barrier stands in the way of harnessing their full potential for high-stakes applications in healthcare: ensuring factual reliability. Specifically, LLMs - both open source and proprietary - are prone to generating hallucinations which are errors arising from: 1) confusion \cite{Jin2022} - low probability tokens are predicted because the model does not know the answer yet responds; 2) confabulation \cite{Ji2023} - the model "thinks" it knows the answer, assigns high probability to a plausible but false statement either due to its factuality or logic; or 3) contamination \cite{Farquhar2024}- the training data contained incorrect, outdated, or contradictory information, which the model has absorbed as truth.

Hallucinations can have significant clinical repercussions. A model suggesting inappropriate discontinuation of potentially life-saving therapy can lead to worse patient outcomes; for instance, non-adherence to standard breast cancer treatment protocols correlates with a 20--30\% reduction in five-year survival \cite{Hershman2011}. Hallucinations is not just a theoretical problem; recent studies have quantified their prevalence in medical LLM outputs. For example, a comprehensive evaluation revealed hallucination rates of 39.6\% for GPT-3.5, 28.6\% for GPT-4.0, and 91.4\% for Bard when generating references for medical systematic reviews \cite{chelli2024hallucination}. As the adoption of LLMs accelerates among clinicians, hospital systems, and the public, so does the urgency to develop robust safeguards against hallucinations \cite{Jin2022,Ji2023,Farquhar2024}.

Multiple strategies have been proposed to mitigate hallucinations in clinical LLMs, yet each faces critical limitations. Fine-tuning on domain-specific corpora aligns models with clinical knowledge and reduces confusion, but offers limited protection against confabulations. Moreover, it inherits risks from contaminated or outdated data \cite{Bender2021} and suffers from catastrophic forgetting when updates overwrite prior knowledge \cite{McCloskey1989,Kirkpatrick2017}. Contamination is a very concerning problem given that these models are trained with data downloaded from the internet and data poisoning attacks as small as $0.001\%$ tokens is enough to make them reproduce harmful medical content \cite{alber2025medical}. Retrieval-augmented generation (RAG) methods, such as REALM , ground outputs in external knowledge bases, improving factuality \cite{Guu2020,Lewis2020}. However, curating and updating these repositories is labor-intensive, prone to bias, and cannot comprehensively reflect the evolving medical landscape. When facts are missing from the databases, hallucinations are exacerbated. Database-free techniques offer complementary solutions. Methods like question resampling with regular-expression filtering \cite{Holtzman2020} and entropy-based analysis \cite{Zellers2019,Kadavath2022} detect confusion by identifying low-confidence, high-entropy outputs. However, they often miss high-confidence hallucinations—i.e., confabulations or contaminations—generated due to nonlinear model behavior or biased training data \cite{Shen2022,Bender2021}.

Recognizing these multifaceted limitations—encompassing contamination and forgetting in fine-tuning, the complexities and gaps in RAG pipelines \cite{barnett2024}, and the partial coverage of database-free checks—we identified an urgent need for transparent, adaptive, and multifaceted solutions. Such solutions must integrate and refine existing mechanisms to ensure continuous factual accuracy in clinical applications. To address this persistent challenge of hallucinations in clinical language models, we introduce CHECK: a novel, dual-pipeline framework that combines structured medical knowledge with probabilistic reasoning to verify factuality at scale.

\section{Algorithm}

\begin{figure}[ht]
    \centering
    \includegraphics[width=1 \textwidth]{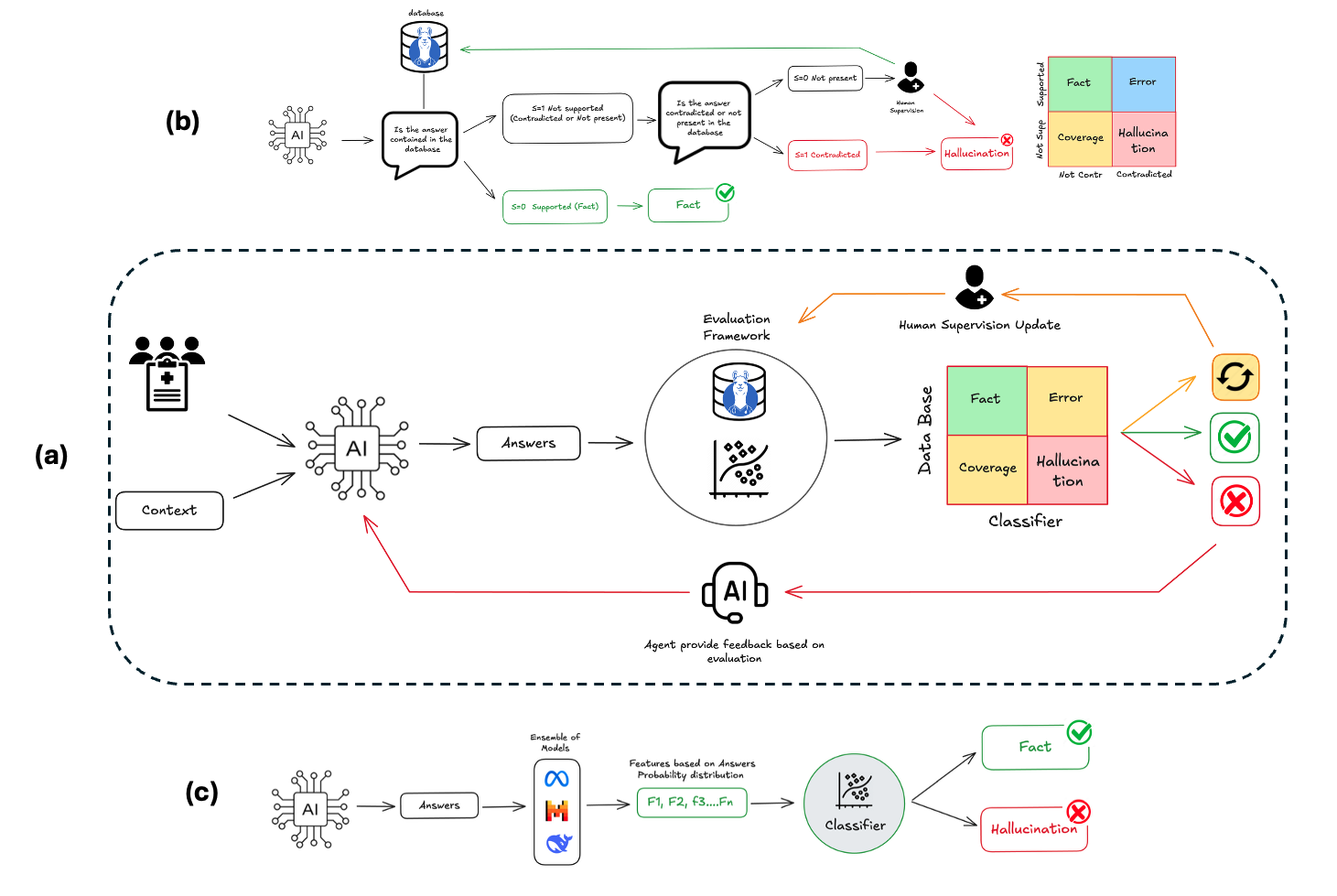}
    \caption{\textbf{(a)} CHECK - Continuous-learning fact-checker framework operationalized at \href{https://thebluescrubs.ai}{The BlueScrubs} platform. \textbf{(b)} Database-driven effective approach against data contamination and adversarial attacks. \textbf{(c)}  Database-Free classification model based on probability distributions over next-token predictions across an ensemble of generative models.}
    \label{check}
\end{figure}

 CHECK (Figure~\ref{check}a) evaluates each {context, query, answer} triplet via two complementary mechanisms: a database-guided pipeline (fig~\ref{check}b) that cross-references model outputs against an open-source, continuously updated, expert-curated clinical knowledge base—open to public scrutiny—and a parallel, model-agnostic classifier (fig~\ref{check}c) that flags hallucinations by identifying statistical signatures of uncertainty and inconsistency across token-level predictions. Grounded in the principle that truth exhibits stability while hallucinations emerge from high-entropy and high-variance distributions across different models or questions, CHECK synthesizes outputs from both pipelines to generate final judgments and escalates unresolved cases to expert review. This architecture not only enables the detection of hallucinations in real time but also supports continuous system refinement through expert feedback and database expansion, offering a transparent and scalable solution for deploying trustworthy medical AI.

\subsection{Stage 1: Database-Driven Fact Checking}
\label{Stage_1}

In the initial evaluation step, CHECK compares the answer against a computationally curated, domain-specific database (as detailed in Section~\ref{sec:database}), leveraging an independent LLM judge employing factual and counterfactual analysis (Figure~\ref{check}b). The model’s output can be evaluated at varying levels of granularity—ranging from entire paragraphs, individual sentences, down to atomic claims—to provide flexibility in identifying inaccuracies or inconsistencies.

For clarity in this analysis, we define the following terms in relation to the curated database content:
\begin{itemize}
\item \textbf{Supported:} The LLM judge can deduce the statement from a set of explicit statements within the database.
\item \textbf{Not Supported:} The LLM judge cannot deduce the statement from any set of explicit statements within the database.
\item \textbf{Contradicted:} The LLM judge can deduce the negation of the statement from a set of explicit statements within the database.
\item \textbf{Not Contradicted:} The LLM judge cannot deduce the negation of the statement from any set of explicit statements within the database.
\end{itemize}
Based on these defined terms, and by analyzing statements and their counterfactuals at each level of granularity, the judge classifies the evaluated content into one of four categories:
\begin{enumerate}
\item \textbf{Fact:} The statement is supported and not contradicted by the database. Score = 1.
\item \textbf{Hallucination:} The statement is contradicted and not supported by the database. Score = 0.
\item \textbf{Judgment Error:} The statement is simultaneously supported and contradicted by the database, implying a potential internal inconsistency within the database itself. Score = 0.5.
\item \textbf{Coverage Gap:} The statement is neither supported nor contradicted by the database. Score = 0.5.
\end{enumerate}

\subsection{Stage 2: Database-Free Hallucination Detection Classifier}

In parallel with database-guided evaluation, CHECK implements a model-agnostic classifier (fig~\ref{check}c) designed to detect hallucinations independently of any knowledge base. Our hypothesis is that, aside from contaminations—which closely mimic factual content and thus require database or expert verification—hallucinated responses exhibit distinctive probabilistic patterns in their token-level predictions if one changes the model or the wording of the question. These patterns reflect the principle that truth is stable, but hallucinations arise from high-entropy and high-variance distributions as the Mark Twain phrase in our title indicates.

To formalize this intuition, CHECK trains a classifier on labeled examples from prior evaluations, using statistical features derived from the next-token probability distributions of multiple LLMs (see Subsection \ref{subsec:fact-vs-hallucination}). Two key information-theoretic features drive this classifier:

\begin{itemize}
\item \textbf{Consistency Across Queries and Models:} Factual answers yield stable next-token distributions even when queries are rephrased or different models are used, while confabulations result in significant variability.
\item \textbf{Distribution Sharpness:} Confused outputs exhibit flatter, near-uniform distributions that indicate uncertainty, whereas factual outputs produce sharply peaked distributions denoting high-confidence predictions.
\end{itemize}

These statistical signatures—rooted in the fact that LLMs are trained to maximize likelihood over factual training data, but not over hallucinations—enable CHECK to detect confusion and confabulation without relying on external databases.

\subsection{Stage 3: Integrated Arbitration and Continuous Feedback} Once both analytic pipelines conclude their evaluations, CHECK integrates their results as follows:
\begin{itemize} 
\item If both the database-driven judgment and the classifier identify the model's answer as factual, the response is confirmed as accurate. 

\item If both analyses identify a hallucination, the output is highlighted as such. Downstream users of CHECK may leverage this information to further tune the LLM, for example by using a 'teacher' model to instruct the original LLM to regenerate the answer with explicit, verified factual context.

\item If the database identifies the response as a hallucination but the classifier deems it factual, the user is alerted, and the model's answer is escalated for human evaluation. This discrepancy indicates potential contamination in the training data where the model has learned an incorrect 'fact' that our statistical classifier, based on output variance, doesn't flag as unusual but the database does - thus making CHECK the first method to protect against data poisoning. 

\item If the database identifies the model's answer as factual but the classifier flags it as a hallucination, the user is alerted. This discrepancy then escalates the text for human evaluation to determine the precise nature of the error, often indicating a possible logic error within the model's reasoning

\item if the database identifies the answer as a coverage gap, the classifier is used as the judge, and the text is flagged for human evaluation to confirm the label as a fact, confusion, confabulation or contamination. 

\end{itemize}

This comprehensive feedback loop continuously refines the curated database, improves the accuracy and coverage of the classifier, and reduces future hallucinations. As a result, CHECK creates an evolving and robust ecosystem where clinical knowledge is systematically verified, enriched, and efficiently deployed, (Figure~\ref{check}a) which has been productionized at \href{https://thebluescrubs.ai}{The BlueScrubs platform}.

\subsection{Medical Use Cases}

To rigorously evaluate CHECK's effectiveness across diverse medical contexts, we applied our framework to four complementary use cases, each representing different aspects of clinical knowledge and reasoning: \\

\textbf{Clinical Trials Benchmark:} Clinical trials represent the gold-standard evidence in medicine, driving clinical standard of care practices and `offering data-driven causal insight into the efficacy, safety, and broader implications of new interventions. As such, they embody the most precise and verifiable definition of factuality in medical applications. To assess CHECK's performance in a realistic and high-impact use case, we developed a benchmark simulating the types of questions physicians ask when interpreting evidence to guide treatment decisions and inform clinical guidelines. Specifically, we constructed a set of 15 clinician-formulated questions (Section S1.2 of the Supplementary Information), covering key trial components such as study purpose, intervention details, eligibility criteria, and both primary and secondary outcomes. These elements reflect the core of evidence-based medicine and serve as a proxy for high-level clinical judgment in knowledge-based decision-making.

\textbf{UMLS Disorders Benchmark:} To test CHECK's ability to generalize beyond clinical trials, we constructed a synthetic benchmark derived from the UMLs Metathesaurus \cite{UMLS}. The UMLS Metathesaurus is a comprehensive and authoritative database of biomedical and health-related concepts, curated by the National Library of Medicine. This benchmark encompasses a wide range of disorders, with each entry including both accurate summaries and synthetically generated hallucinations related to definitions, pathophysiology, risk factors, diagnostic criteria, and treatment strategies. This dataset simulates structured diagnostic reasoning and allows us to evaluate whether the statistical signatures of hallucination identified in clinical trials extend to a broader spectrum of medical concepts (details in Section S1.3 of the Supplementary Information). The task mirrors real-world scenarios in which clinicians construct differential diagnoses or choose among therapeutic options, making it a valuable test of CHECK’s diagnostic reasoning support.

\textbf{MedQA (USMLE) Benchmark:} The MedQA (USMLE) benchmark \cite{app11146421}  - the United States Medical Licensing Examination - reflects clinical reasoning under uncertainty. Each question requires integrating foundational knowledge, pathophysiology, clinical guidelines, and diagnostic strategy—skills directly transferable to real-time understanding in patient care. This dataset enables detection of hallucinated rationales or logic failures within the context of complex clinical reasoning. Although its clinical utility has been questioned, this benchmark has become a gold standard for evaluating AI in medical education and practice and offers a surrogate to evaluate how models perform in medical reasoning tasks. 

\textbf{HealthBench Benchmark:} While the MedQA (USMLE) benchmark assesses clinical reasoning in a structured examination format, the recently introduced HealthBench benchmark \cite{healthbench} directly evaluates an LLM's capacity for safe and effective communication in realistic, multi-turn clinical conversations. These dialogues, designed to simulate interactions with both patients and clinicians, require nuanced understanding, contextual awareness, and the generation of accurate, empathetic, and clinically appropriate responses. Given that critical medical errors often stem from communication breakdowns, this benchmark provides a crucial testing ground for CHECK's ability to identify hallucinations within complex, fluid conversational exchanges. Evaluating our framework on HealthBench allows us to determine if the statistical signatures of hallucination identified in more structured settings translate to the dynamic and open-ended nature of real-world clinical discourse, thereby establishing CHECK's utility in ensuring trustworthy AI for direct patient and clinician support.

\section{Results}

We assessed the factual accuracy of LLama3.3-70B-Instruct—a state-of-the-art open-source model—by evaluating its responses to a total of 15 clinically relevant questions. These questions pertained to content from a collection of 100 clinical trials, randomly selected from our database of 68,000 processed trials (all trials with reported results as of November 2024 in \href{https://clinicaltrials.org}{ClinicalTrials.org}). This evaluation used our database-driven fact-checking approach (Section~\ref{Stage_1}) to systematically identify factual statements and hallucinations.

\begin{figure}[ht]
    \centering
    \includegraphics[width=0.85\textwidth]{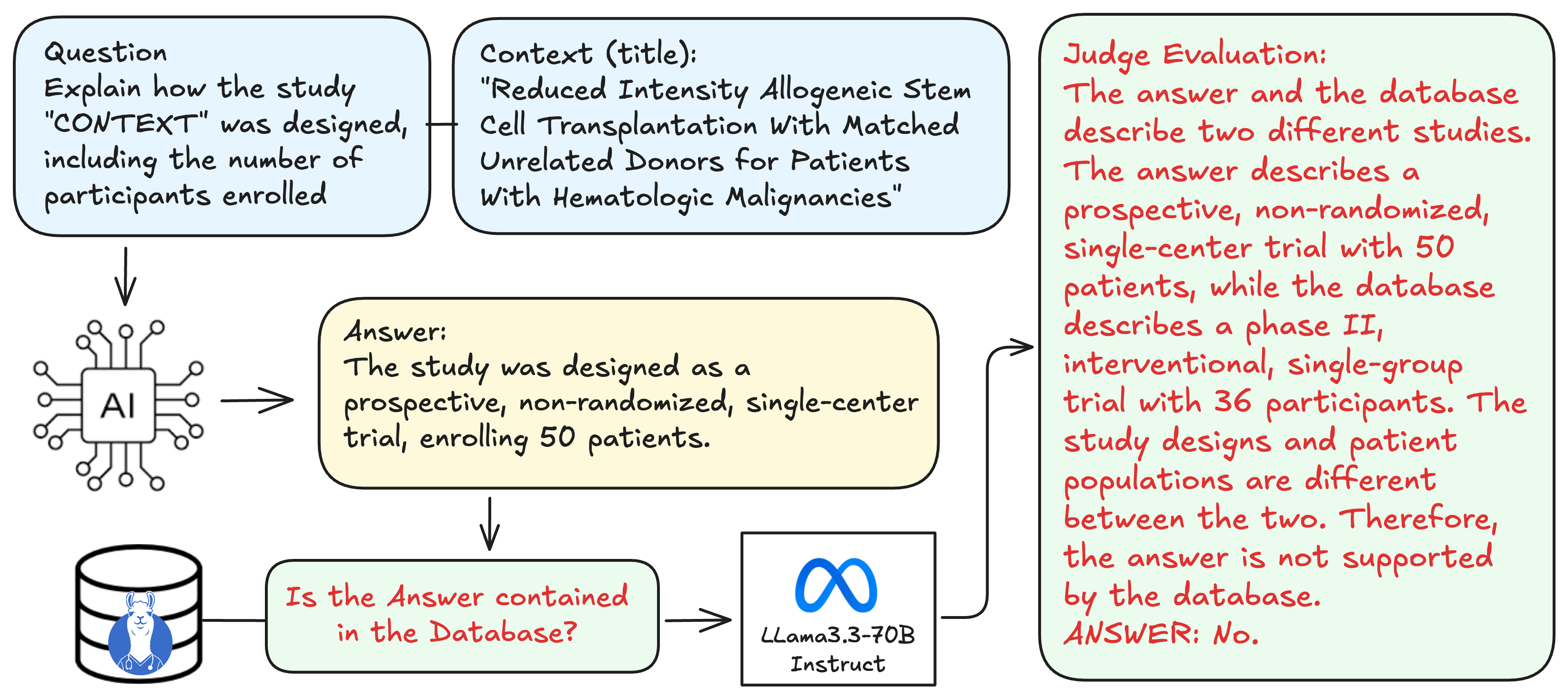}
    \caption{Factual analysis performed by an LLM judge on model-generated answers about clinical trial NCT00818961 (Donor Stem Cell Transplant in Treating Patients With High-Risk Hematologic Cancer). The evaluation can be verified using trial description and recruitment information section from \href{https://clinicaltrials.gov/study/NCT00818961?tab=table}{ClinicalTrials.gov}}
    \label{judge}
\end{figure}

We selected this open-source LLM to support scientific reproducibility, as results can be corroborated over time - a fundamental difference with proprietary models - altough we also study priopretary models.  For each trial, the model received one of three context types: (1) title-only, representing minimal contextual support similar to zero-shot inference; (2) complete raw JSON trial records downloaded from clinicaltrial.org, simulating conventional retrieval-augmented generation (RAG); or (3) our structured clinical summaries curated specifically for this study through our computational pipeline (described below). To determine answer correctness, we employed LLama3.3-70B-Instruct as an independent judge performing dual evaluations: first assessing whether the database directly supported the answer (factual analysis), then determining whether the database contradicted it (counterfactual analysis). For both analyses, the judge provided binary decisions with explanatory justifications. Figure~\ref{judge} illustrates this database-driven approach with an example of hallucination detection. Additional examples of factual and hallucinated responses are provided in Section S2.2 of the Supplementary Information.

\subsection{Database Hallucination Evaluation}

\begin{figure}[ht]
    \centering
    \includegraphics[width=1\textwidth]{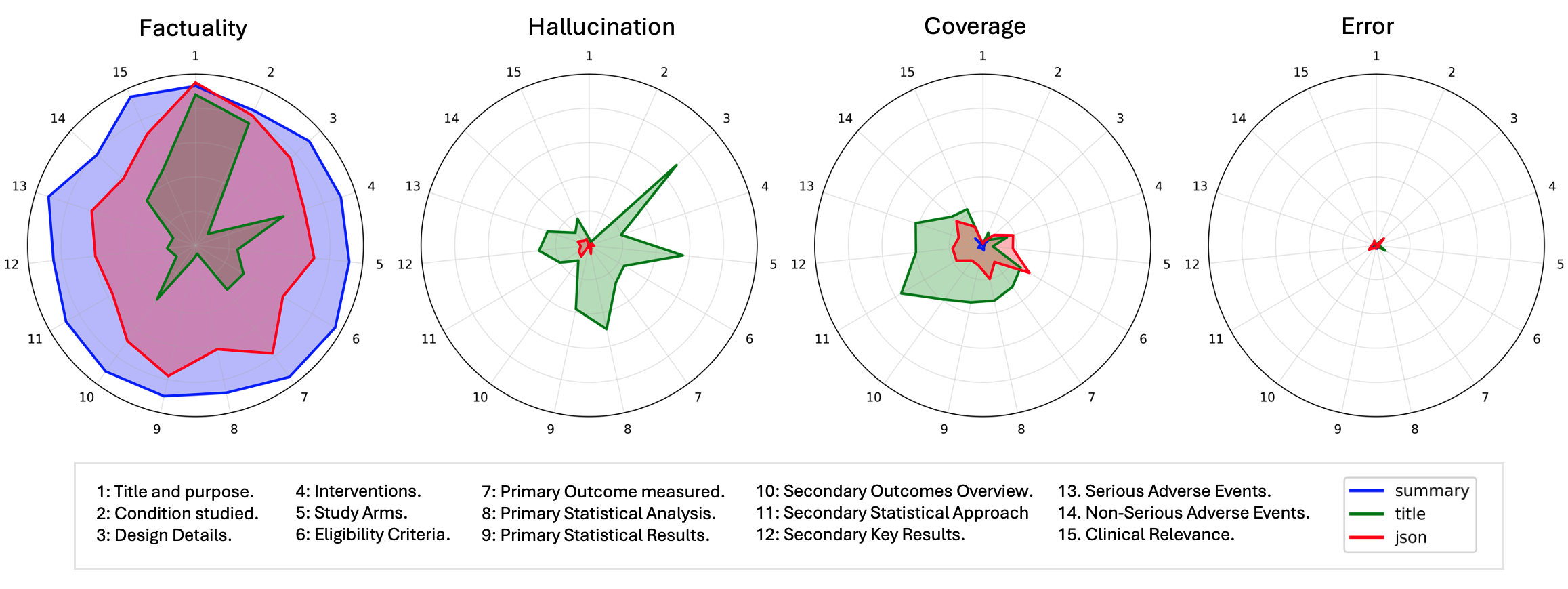}
    \caption{Distribution of LLM Answer Labels Across 15 Question Categories, by Context Source. This radar chart illustrates the average percentage of responses falling into 'Factuality', 'Hallucination', 'Coverage', and 'Error' categories, evaluated for 15 distinct question types (labeled 1-15 in the legend). Each radial grid line represents a 25\% increment for the given category. Responses were generated by the LLM using three different context sources: 'summary' (blue), 'title' (green), and 'json' (red). As clearly demonstrated, the model answering with the curated summaries as context consistently achieves the highest factuality and a significantly lower rate of hallucinations compared to using only the trial title or raw JSON data.}
    \label{radar}
\end{figure}

Figure~\ref{radar} and Table~\ref{table-1} summarize the performance of the LLama3.3-70B-Instruct model \cite{llama} in three input contexts for the clinical trial questions: trial title, full JSON file, and curated clinical summaries. The \emph{summary} context achieved the highest factual accuracy, with 97\% of responses labeled as factual and only 0.3\% as hallucinations—a 100-fold reduction compared to the \emph{title} context (31\% hallucinations). While the JSON input offered broader informational coverage, its nested structure occasionally hindered factual precision (more detailed in Section S2.1 of the Supplementary Information). Notably, hallucinations were most frequent when the model received only the trial title (31\%) to respond, particularly for questions probing study design or outcomes, where insufficient context led to fabricated details - in line with other studies \cite{eval-harness}. Coverage gaps, where answers were plausible but unsupported, were more prevalent in the title (29\%) and JSON (16\%) conditions. These "coverage" labels signify content that our database-driven judge could neither explicitly support nor contradict. While gaps arising from JSON input appeared more likely to represent factual information not yet captured by our database's granularity, those from the title context seemed more likely to be hallucinations, given the high hallucination rate observed with minimal context. This highlights a key limitation of relying solely on database-driven fact-checking, providing a perfect example of how combining independent techniques can enhance automated decision-making in complex evaluation scenarios. The exact nature of these coverage gaps, and how they are resolved through our integrated approach, is addressed in a subsequent analysis. Judge inconsistency (error) remained rare ($<$3\%) across all settings, confirming high internal reliability. These results also demonstrate that structured, human-curated summaries most effectively ground LLM outputs in verifiable clinical evidence, validating CHECK's layered verification approach for minimizing risk in high-stakes domains.

\begin{table}[ht]
\label{table1}
\centering
\begin{tabular}{lcccc}
\toprule
Context source & Factuality & Hallucinations & Coverage & Error \\
\midrule
title & 38\% & 31\% & 29\% & 2\% \\
json & 78\% & 3.6\% & 16\% & 2.4\% \\
summary & 97\% & 0.3\% & 2\% & 0.7\% \\
\bottomrule
\end{tabular}
\caption{Label distribution (in percentage) of 1500 evaluated answers generated by LLama3.3-70B-Instruct, for each input context source (Title, JSON, Structured Summary). The dataset consists of 1500 question--answer pairs derived from 15 questions applied to 100 clinical trials.}
\label{table-1}
\end{table}

\subsection{Database-Free Hallucination Detection}
We developed a stacking classifier, trained on the labeled dataset derived directly from our database-driven fact-checking stage. This dataset comprised approximately 80\% factual and 20\% hallucinated examples, with labels assigned based on the outcomes of factual and counterfactual analysis of LLM answers to clinical trial questions. To create clear and distinct training signals for the classifier, we strategically selected data from model responses produced using high-quality context: \emph{summaries} (which predominantly yielded factual content) and \emph{titles} (which were identified as significant sources of hallucination; see Table~\ref{table-1}). Details on the extracted features and classifier training are provided in the Methods (Section 5.4) and Supplementary Information (Section S2.2). The classifier achieved excellent performance with an AUC of 0.95 on this clinical trial data, demonstrating robust discrimination between factual and hallucinated content. Overall accuracy reached 91\%, significantly outperforming both random and majority-class baselines. Detailed performance metrics, including precision, recall, and F1-scores for both classes, are also provided in the supplementary materials.

\subsection{Integrated Arbitration and Continuous Feedback Application}

To demonstrate the practical utility of CHECK’s arbitration mechanism (Section 2.5, Stage 3), we evaluated its performance in real-world scenarios where factuality could not be definitively established through database coverage alone. These cases, labeled as “coverage gaps” by the database-driven pipeline, were adjudicated by CHECK’s independent, database-free classifier, which assigned a hallucination probability to each response (per Section 2.3). This enabled ranking of ambiguous outputs by their likelihood of being hallucinated, even in the absence of a definitive ground truth.

To validate this probabilistic arbitration, we conducted a targeted human review of the 10 responses with the highest predicted hallucination probabilities and the 10 with the lowest. Only one misclassification was identified among the 20 cases, yielding a 95\% agreement between the classifier and human judgment. High-probability cases often contained subtle inaccuracies or unsupported claims—plausible yet fabricated content not directly contradicted by the database—highlighting classic LLM confabulations. These hallucinations predominantly arose when the input context was minimal (e.g., only a trial title), consistent with prior findings that LLMs hallucinate more when relying on sparse prompts.

Conversely, responses assigned low hallucination probabilities were typically judged factual by experts. Notably, many of these “coverage gap” cases reflected true information overlooked by the LLM judge due to misinterpretation of rephrased or nuanced content. This review underscores the value of CHECK’s dual-pipeline design: the database-free classifier reliably detects hallucinations when database confirmation is unavailable, enhancing both the precision and robustness of the overall system in high-stakes clinical applications.

\subsection{Generalization capabilities on UMLS disorders benchmark}

Remarkably, when applied to the entirely different UMLS disorders dataset (UMLS disorders case study), the classifier maintained its robust performance (AUC = 0.96), with balanced results across both factual and hallucinated content. This cross-domain generalization demonstrates that our database-free approach captures fundamental statistical signatures of hallucination that transcend specific medical domains, offering a reliable safeguard against misinformation in diverse clinical contexts.

\subsection{Feature Importance and Hallucination phenotype}

Feature importance analysis revealed two key categories of statistical predictors distinguishing facts from hallucinations: token-level uncertainty metrics and ensemble distribution divergence measures (see Fig S8). Token uncertainty features—including entropy and log probabilities of low-ranking tokens—effectively identified confusions, which typically exhibit higher entropy and flatter probability distributions, indicating model uncertainty when generating content beyond its confident knowledge. Complementing this, Kullback-Leibler divergence between probability distributions across our model ensemble (LLama3.1-8B-Instruct, LLama3.1-70B-Instruct, LLama3.3-70B-Instruct, Nemotron-70B \cite{llama}, DeepSeek-V1 \cite{deepseek}) proved a highly important independet predictor; effectively detecting confabulations (one model sees a statemetn with high probability but the others do not). This dual approach, which leverages both internal model uncertainty and cross-model agreement, enables our classifier to achieve robust hallucination detection by capturing the fundamental statistical signatures that differentiate factual from fabricated content and validate our hypothesis. Detailed feature importance analysis and principal component analysis are provided in the Figure S5 of the supplementary materials.

\subsection{Clinical Reasoning: MedQA (USMLE) Benchmark}

Building on our success with factual verification in clinical trials and UMLS disorders, we extended CHECK to the more complex domain of clinical reasoning using the MedQA (USMLE) benchmark \cite{app11146421}. This multiple-choice question dataset presents a different challenge: distinguishing between factual (correct) and hallucinated (incorrect) answer choices in a high-stakes medical examination context that also include medical reasoning. Our classifier, using the same statistical features derived from our model ensemble, maintained robust performance with an AUC of 0.95 for both fact and hallucination classes.

We further leveraged the classifier to identify correct answers by examining the fact probability assigned to each of the four choices per question. In particular, by selecting the answer with the highest probability of facts, we achieved a Top-1 accuracy of 79\% in the MedQA test set - significantly outperforming the baseline 71\% obtained when computing performance based on average predictions of our ensemble under harness standard evaluation\cite{eval-harness}. The Top-2 accuracy reached 93\%, demonstrating the classifier's strong ranking capability for clinical reasoning tasks.

\subsubsection{Hallucination Probability as a Reliability Indicator}

To demonstrate CHECK's practical utility beyond binary classification, we explored its capacity to rank answers by hallucination probability, enabling targeted interventions for challenging questions. This approach allows for dynamic resource allocation—escalating to human experts or increasing computational resources only when necessary.\

\begin{figure}[ht]
    \centering
    \includegraphics[width=0.7\textwidth]{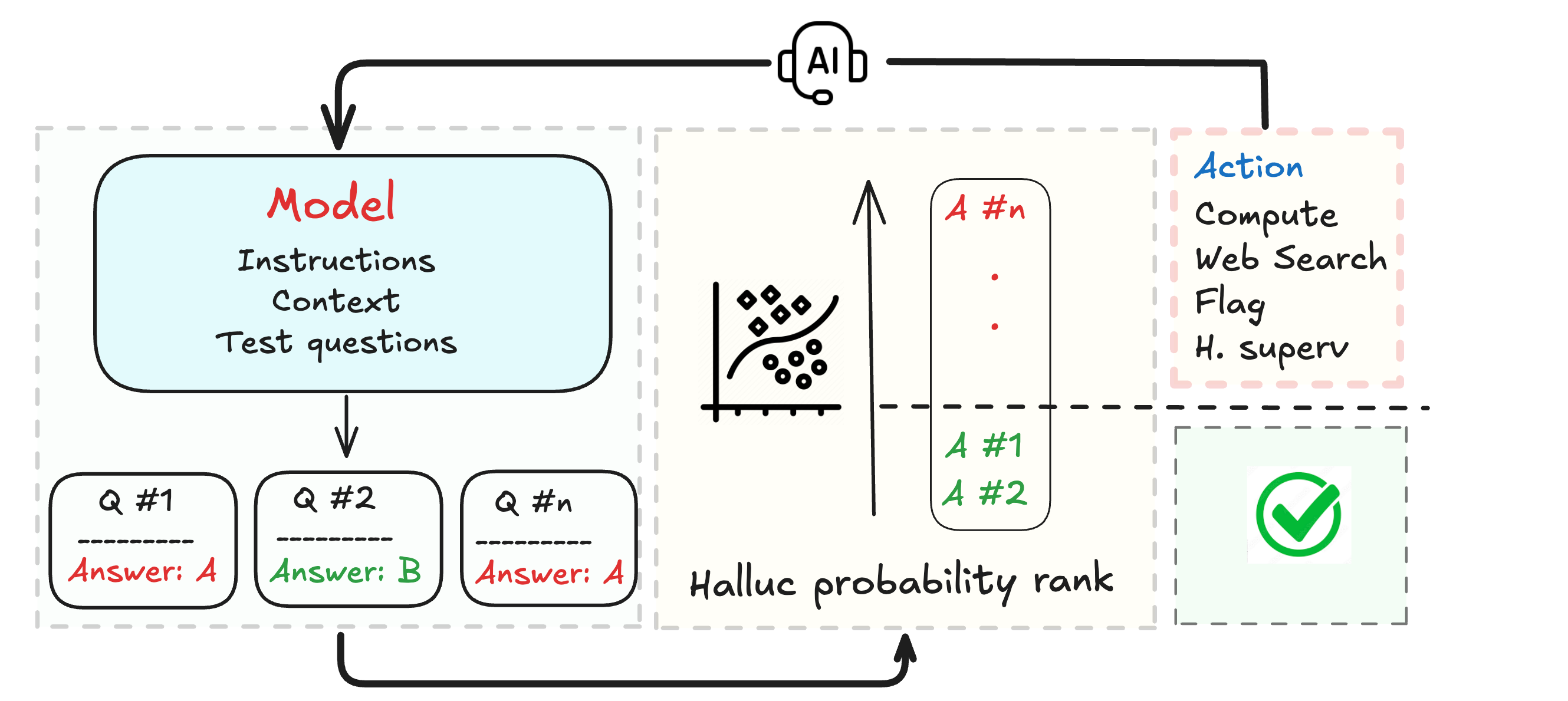}
    \caption{Adaptive intervention framework based on hallucination probability ranking. Responses are first ranked by their hallucination probability scores from CHECK's classifier. Based on a predetermined threshold, high-risk responses trigger specific interventions: increased computation time, web-based verification, automated flagging, or human expert review. Lower-risk responses are accepted directly. The intervention agent can provide feedback to the original model, creating a dynamic improvement loop. This selective approach enables efficient resource allocation by focusing additional verification steps only on responses with higher hallucination risk.}
    \label{percentiles}
\end{figure}

We established baseline performance using GPT-4.o, which achieved 87.67\% accuracy on the MedQA (USMLE) test set. Adding Chain-of-Thought (CoT) reasoning improved accuracy to 90.89\% (Table~\ref{benchmark}). For each question, we then used our classifier to assign hallucination probability scores to the predicted answers in both configurations.

\begin{figure}[ht]
    \centering
    \includegraphics[width=0.6\textwidth]{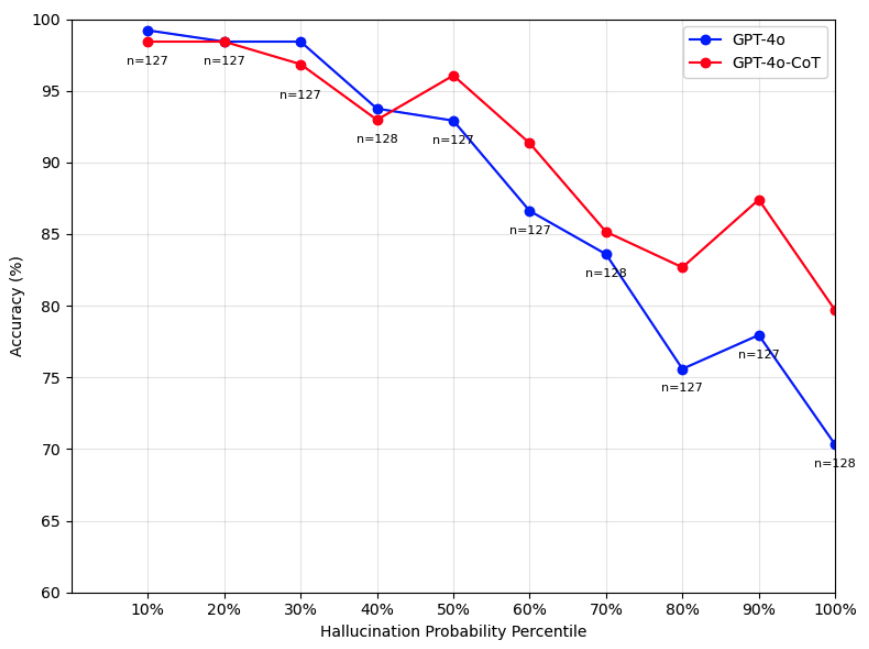}
    \caption{Model Performance across Hallucination Probability Percentiles. Accuracy of GPT-4o baseline and Chain-of-Thought (CoT) on the MedQA test set, evaluated within 10 percentile bins ranked by hallucination probability. Lower percentiles correspond to lower predicted hallucination probability.}
    \label{percentiles-rank}
\end{figure}

Organizing the predictions into 10 percentile bins based on hallucination probability revealed a clear pattern (Figure~\ref{percentiles-rank}). In the lowest 30\% of hallucination probabilities, both models achieved exceptional accuracy ($>96\%$). As hallucination probability increased, performance gradually declined to 86-96\%. In the highest 40\% of hallucination probabilities , accuracy dropped more significantly to 70-87\%.

This strong correlation between predicted hallucination probability and model performance demonstrates CHECK's effectiveness as a reliability indicator. Leveraging this insight, we selectively increased computational resources by running 12 iterations of CoT reasoning only on the top 40\% of questions with highest hallucination probability. This targeted approach achieved 92.06\% accuracy, surpassing the previous state-of-the-art result of 91.1\% from Med-Gemini-L 1.0\cite{saab2024capabilitiesgeminimodelsmedicine} while using computational resources more efficiently \cite{snell2025scaling}. In particular, a recent analysis of the MedQA data set by attending clinicians \cite{saab2024capabilitiesgeminimodelsmedicine} reveals that approximately 4\% of the questions contain missing information, and an additional 3\% potentially have labeling errors. This suggests that the achievable upper bound on this benchmark may be closer to our reported accuracy, highlighting the significance of our approach in light of the dataset's inherent limitations.  Furthermore, our targeted allocation of computational resources achieves state-of-the-art performance at a fraction of the operational cost compared to approaches that rely on external web search\cite{saab2024capabilitiesgeminimodelsmedicine} or apply increased test-time compute\cite{snell2025scaling} across the entire dataset.

\begin{table}[ht]
\centering
\begin{tabular}{lcc}
\toprule
\textbf{Model} & \textbf{Accuracy (\%)} & \textbf{Cost} \\
\midrule
GPT-4o Base & 87.7 & Low \\
GPT-4o CoT & 90.9 & Medium \\
\textbf{GPT-4o CoT + CHECK (12x, top 40\%)} & \textbf{92.1} & Medium \\
\midrule
Med-Gemini (previous SoTA) & 91.1 & High \\
\bottomrule
\end{tabular}
\caption{\textbf{Performance Comparison on MedQA (USMLE) benchmark.} Accuracy (\%) of GPT-4o across different strategies: standard prompting (Base), Chain-of-Thought reasoning (CoT), and targeted 12-iteration majority voting applied only to the 40\% of questions with highest hallucination probability (CoT 12x, top 40\%). The bottom row shows the previous state-of-the-art result from Med-Gemini-L 1.0\cite{saab2024capabilitiesgeminimodelsmedicine}. Cost indicates relative computational expense of each approach.}
\label{benchmark}
\end{table}

\subsection{HealthBench: Generalization to Realistic Clinical Conversations}

While our previous evaluations demonstrated CHECK's effectiveness across structured clinical trials, generalized medical concepts (UMLS disorders), and high-stakes medical examinations (USMLE MedQA), we further extended our assessment to evaluate its robustness in the context of realistic, multi-turn clinical conversations. For this, we utilized the newly introduced HealthBench benchmark \cite{healthbench}. This benchmark presents a critical challenge for AI in healthcare by demanding nuanced understanding and the generation of accurate, safe, and contextually appropriate responses in scenarios simulating interactions with both patients and clinicians. The inherent complexity and open-ended nature of these dialogues make hallucination detection particularly critical, as errors could directly impact clinical decision-making.

To rigorously assess our framework within this conversational domain, we constructed an evaluation dataset from the HealthBench Consensus and HealthBench Hard subsets. We first generated baseline responses using GPT-4o, leveraging the github pipeline released by Open AI \cite{simpleevals}. To create a challenging set of hallucinated examples, we employed an adversarial strategy: we prompted GPT-4o to rewrite each original factual response as a hallucination, meticulously preserving its original structure and stylistic elements (details in section (\ref{umls}) and Figure S1). This ensured that the fabricated hallucinations closely mimicked legitimate outputs, making them a significant challenge for detection.

\subsubsection{Classifier Training with GPT-4o Statistical Features}
To apply the core principle of our detection method—measuring variance across token-level predictions—using the GPT-4o API instead of an ensemble of models, we introduced linguistic variations to the questions by rephrasing each original question five times but keeping the context constant. Subsequently, for each original and fabricated text, we prompted GPT-4o to classify its content (paired with each of the five rephrased questions) as either containing a hallucination or not, specifically by yielding a single-token "yes" or "no" answer. From these controlled generation steps, we extracted log-probabilities for the top 20 predicted tokens at each generation step. These served as the foundation for our statistical features, consistent with our approach in previous benchmarks (see Subsection \ref{subsec:fact-vs-hallucination}). A robust baseline accuracy of 96.2\% for both factual and hallucinated classes was established by taking the mode of GPT-4o's five answers on the test set, highlighting the model's strong inherent ability to self-identify hallucinations. Our classifier, trained on these meticulously engineered statistical features, demonstrated superior performance, achieving an accuracy of 97.7\%.

\subsubsection{Classifier Training with Ensemble Model Statistical Features}
We then repeated the database-free hallucination detection approach using an ensemble of open-source Llama models (Llama3.1-8B-Instruct, Llama3.1-70B-Instruct, Llama3.3-70B-Instruct). For this, we performed forced inference of the full text (akin to a harness evaluation), providing the original question and the GPT-4o generated answer to each component of the ensemble. We then built statistical features extracted from their token probability distributions. This multi-model ensemble approach yielded an even more impressive accuracy of 98.4\%. Please note that these models independetly perform significantly worse in these questions as compared to ChatGTP 4.0. These results, however, further underscoring the power of measuring variance across diverse models for robust hallucination detection even when they perform worse than the original model that generated the answers.

\subsubsection{Ranking Responses by Hallucination Probability for Confidence Assessment}
\begin{figure}[ht]
    \centering
    \includegraphics[width=0.55\textwidth]{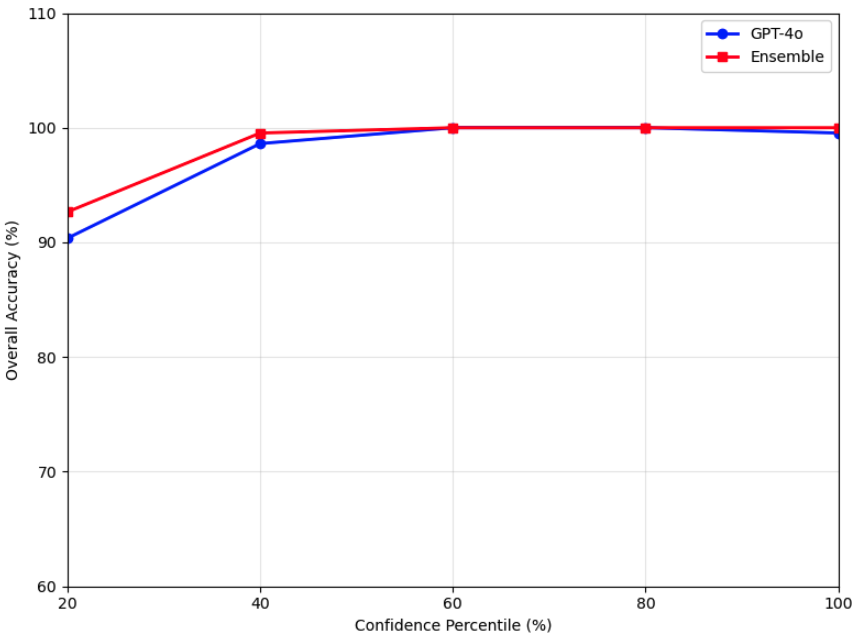}
    \caption{Overall Model Performance across Confidence Percentiles. Accuracy of GPT-4o (using rephrased questions) and the Llama model ensemble approach on the HealthBench benchmark, evaluated within 20 percentile bins ranked by model prediction confidence. Lower percentiles correspond to low model prediction confidence.}
    \label{percentiles_healthbench}
\end{figure}

A core strength of CHECK is its ability to identify clear confidence intervals for model predictions, offering a crucial reliability indicator for taking further actions and enhancing precision when needed. As depicted in Figure \ref{percentiles_healthbench}, when predictions are organized by their probability values, CHECK consistently achieves nearly 100\% accuracy for both classes (hallucinations and facts) above the 20\% baseline. Conversely, the majority of errors are concentrated in the lower 20\% confidence percentile, where the model assigns lower probability to its predictions. CHECK proper calibration is of paramount importance for identifying instances where the model is less certain, thereby facilitating targeted human oversight or adaptive computational interventions, as discussed in Section 2.5.1.

\section{Discussion}

CHECK introduces a hybrid, self-improving framework that substantially advances factual reliability in clinical language models. By combining structured medical databases with a model-agnostic, information-theoretic classifier, CHECK reduces hallucination rates by over 100-fold ( $< 0.3\%$) on clinical trial questions—outperforming established safety thresholds for high-risk medical processes such as drug administration errors (5–10\%) \cite{MacDowell2021}; therefore paving the way for the wide incorporation of LLMs in clinical practice.

The framework’s effectiveness was demonstrated across three clinically relevant tasks that span the continuum of evidence-based practice: from knowledge retrieval and synthesis (clinical trials) and structured diagnostic reasoning (UMLS disorders), to medical reasoning under uncertainty (USMLE MedQA) or multi-turn realistic medical conversations (HealthBench). The consistency of CHECK’s performance across them—notably, AUCs of 0.95–0.96 for hallucination detection even without a database—highlights the generalizability of its core insight: factual statements produce stable, low-entropy token distributions, while hallucinations diverge in structure and confidence across queries and models. Beyond detection, CHECK also acts as a reliability signal and orchestration engine. On MedQA (USMLE), hallucination probabilities enabled targeted test-compute escalation, improving GPT-4o’s accuracy to 92.1\%—a new state-of-the-art—while optimizing computational cost. This indicates that CHECK can prioritize human review or enhanced prompting only when truly needed, offering a path toward more efficient and safer AI deployment in practice.

Perhaps most importantly, CHECK is not a static verifier but a continuously learning system. Discrepancies between its database-anchored and classifier-driven evaluations trigger expert review,  and validated insights are reintegrated into both components—expanding coverage and refining detection capabilities over time. Central to this architecture is a commitment to transparency: the clinical knowledge base supporting CHECK is open-source and publicly auditable, allowing for community validation, rapid updates, and reduced risk of silent failure. This principle is operationalized through \href{https://thebluescrubs.ai}{The BlueScrubs} platform, which maintains curated, versioned clinical datasets and supports collaborative model oversight. By surfacing and correcting hallucinations—whether stemming from knowledge gaps, logical errors, or contaminated training data—CHECK offers not only improved performance, but a reproducible and extensible foundation for medical AI governance. Its dual-mode architecture is broadly compatible with clinical decision support, quality assurance, medical education, and evidence synthesis, advancing a blueprint for trustworthy, self-correcting AI in high-stakes domains.

Critically, CHECK’s hallucination detection capabilities remain effective even when evaluating outputs from models significantly more powerful than those in its own ensemble. For instance, GPT-4o consistently outperforms individual LLaMA models on tasks like HealthBench; yet CHECK’s ensemble reliably flags hallucinations in GPT-4o’s outputs—despite having no access to its internal weights or training data. This underscores a key strength of CHECK: its ability to exploit general statistical properties of hallucinations, particularly elevated variance in token-level probability distributions across independently trained models when confronted with hard or underspecified questions. These variance patterns are not contingent on how a given answer was generated, making CHECK a genuinely model-agnostic tool. By detecting distributional instability as a proxy for factual unreliability, CHECK provides an additional factuality safeguard—independent of the underlying generation model—and thereby enhances trust in clinical applications of even the most advanced LLMs.

\section{Methods}

\subsection{Database Curation Pipeline.}

Raw clinical trial files, often in complex and heterogeneous formats like JSON, contain deeply nested structures, highly technical descriptions and regulatory language that hinder automated parsing and interpretation, even for advanced language models. These files are fragmented and technically dense,  making automatic accurate data extraction or verification challenging. In contrast, a structured clinical trial summary database has the potential to distill key findings, clinical relevance, and risks into an interpretable format, enhancing usability for clinicians, researchers, and AI agents. 

\label{sec:database}

To build a concise, clinically oriented database from raw trial files, we implemented a multi-step approach that isolated key sections from the JSON data downloaded from \href{https:\\clinicaltrials.org}{ClinicalTrials.gov} (519,600 trials as of November 15, 2024). Each trial file comprises three main keys:

\begin{itemize} \item \textit{Protocol Section:} Contains study identification, design, and eligibility details. \item \textit{Results Section:} Encompasses participant flow, outcome measures, and adverse events. \item \textit{Derived Section:} Maps conditions and interventions to term identifiers and names, indicating their clinical relevance. \end{itemize}

\noindent In the first stage, we focused on the \textit{Protocol Section} to generate a structured \textit{Overview} of each trial, highlighting its purpose, design, interventions, and recruitment criteria. Rather than processing entire JSON files at once, we prompted a state-of-the-art open-source language model (Llama3.3-70B-Instructed) to transform only the relevant protocol data into a reasoning-oriented summary, thereby reducing complexity and improving the clinical coherence of the resulting text. The specific prompt structure and examples of generated text are provided in the supplementary materials.

\noindent Next, to examine actual study findings, we filtered the dataset to include only the 68,152 studies with reported results. We then repeated our pipeline approach for the \textit{Results Section}, prompting the same language model to extract and interpret primary and secondary outcomes as well as adverse events. By limiting each generation step to a well-defined subset of the original data, this process yields more focused and interpretable summaries, clarifying the statistical methods employed and the clinical significance of any observed events. The prompting details and sample outputs for this stage are also included in the supplementary materials.

\noindent Finally, information from the \textit{Derived Section} was integrated to map clinical conditions and interventions to standardized terms and identifiers, ensuring that the curated database preserved both semantic clarity and clinical context. This structured repository of trial overviews and result summaries, enriched by domain-specific reasoning cues, provides an accessible and authoritative knowledge base. By systematically breaking down each trial into distinct parts, our pipeline effectively addresses the challenges posed by the complex and heterogeneous nature of raw JSON files, facilitating subsequent analyses and reliable fact-checking for large language models in clinical applications.

\subsubsection{Human Review of Curated Summaries} To further validate the utility and accuracy of the curated clinical trials database, a subset of generated summaries was subjected to human evaluation. Each evaluator received the following instructions and format for assessment, which emphasizes both completeness and factuality:

\begin{itemize} \item \textit{Completeness Score (1--5):} Measures how thoroughly the answer captures the scope of information found in the original trial source, with 1 denoting minimal coverage and 5 denoting comprehensive representation. \item \textit{Factuality Score (1--5):} Evaluates how accurately the provided information aligns with the original trial record, independently of completeness. Even a concise answer may receive a high factuality score if all stated details match the official record. \end{itemize}

Evaluators were given excerpts from the curated database, each linked directly to its corresponding entry on ClinicalTrials.gov. They then assigned completeness and factuality scores based on a comparison of the summary content against the official trial documentation. This approach ensured that the resulting evaluations captured both the breadth of the extracted information (completeness) and its correctness (factuality). Full details of the human evaluation methodology and results are provided in the Supplementary Information.

\subsection{Validation and Evaluation of Clinical Language Models Fact Verification} 
\label{sec:counterfactual}

To assess the reliability of the LLM judge, we conducted a human review of a representative subset of its evaluations. In particular, 3 human reviewers - PhD Scientists - compared each statement generated by the primary LLM to the corresponding judge's verdict (e.g., “supported” or “contradicted”) relative to the curated summary database. Reviewers examined whether the judge correctly understood the question, applied the factual or counterfactual criteria, and ultimately reached a conclusion that aligned with the consensus of human evaluation. This process not only validated the accuracy of the judge’s classification but also provided insights into potential areas where the judge might misinterpret or overlook salient details.

\subsection{Developing a Classifier for Fact vs.\ hallucination} 
\label{subsec:fact-vs-hallucination}

Building on the counterfactual pipeline described above, we collected labeled data by systematically evaluating each LLM-generated response against the curated clinical trials database. When the LLM judge performed its factual/counterfactual analysis, every evaluated answer was assigned one of four outcomes (factual, hallucination, coverage gap or error) for different levels of granularity. Focusing on the statements labeled as either factual or hallucination, we  built a labeled dataset of \textit{fact} vs.\ \textit{hallucination}. 

\subsection{Feature creations}
\label{subssec:feature-creation}
We wanted to characterize each statement at different levels of granularity. We processed each LLM-generated response using \textit{forced inference}. We sequentially collect the logarithm of the probability (common output of inference engines) of each token in the dictionary - including the one generated - conditioned on all previously generated tokens, across an ensemble of five distinct language models (LLama3.1-8B-Instruct, LLama3.1-70B-Instruct, LLama3.3-70B-Instruct, Nemotron-70B, DeepSeek-V1) \cite{llama,deepseek}.

\begin{figure}[ht]
    \centering
    \includegraphics[width=1\textwidth]{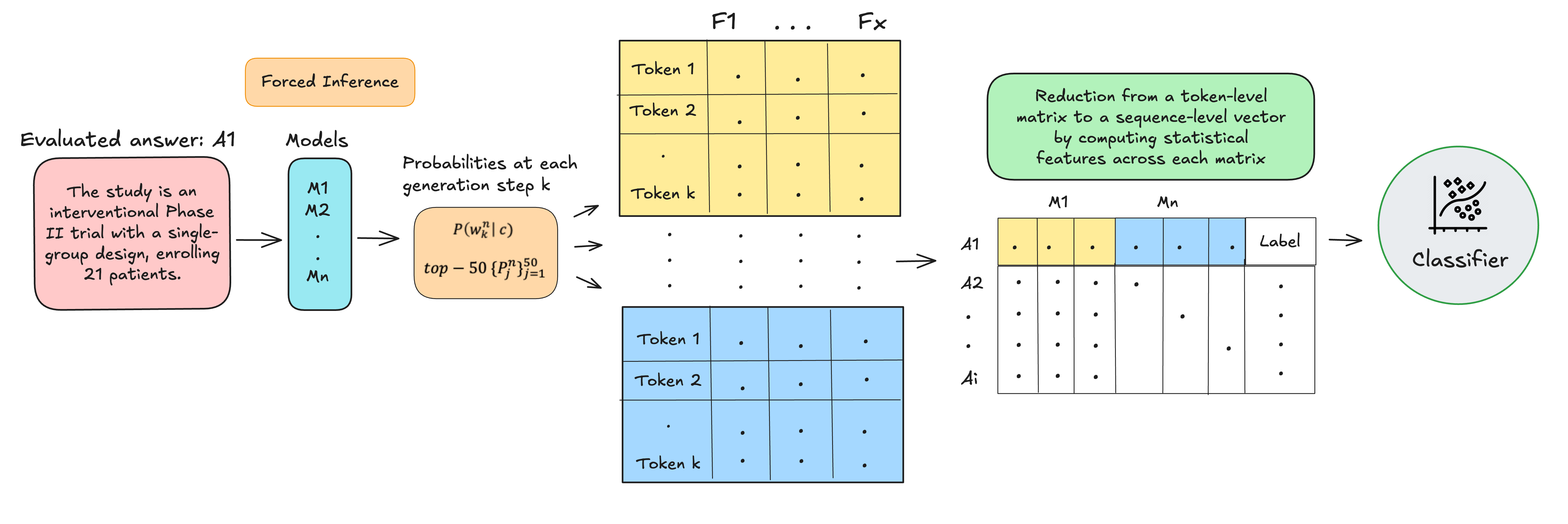}
    \caption{Feature extraction pipeline for factuality/hallucination detection. The process involves forced inference of each answer through an ensemble of language models, generating token-level probability distributions that are transformed into statistical features. These features capture both independent token statistics and cross-model interaction patterns, ultimately producing a compact representation for classification.}
    \label{fig:features}
\end{figure}

This approach allowed us to quantify precisely how each model assessed the likelihood of the observed sequence of tokens. As depicted in Fig~\ref{fig:features} from these token-level probability distributions, we constructed independent statistical features characterizing it—including entropy (equation~\ref{H}), generated-token rank, and probability of the top-50 predicted tokens at each generation step—as well as interaction features, such as pairwise Kullback–Leibler (KL) divergences (equation~\ref{kl}) between the probability distributions from different models or restating questions. Finally, we reduced each resulting feature matrix (rows given by tokens and columns features) at the granularity level by extracting statistical descriptors up to the fifth statistical moment (mean, variance, skewness, kurtosis, and hyperskewness), thereby obtaining a compact yet informative feature representation for each generated answer together with its label.

\begin{equation}
\label{H}
H(X) = -\sum_{i=1}^{n} p(x_i) \log p(x_i)
\end{equation}

\begin{equation}
\label{kl}
D_{KL}(P\|Q) = \sum_{i=1}^{n} P(x_i) \log \frac{P(x_i)}{Q(x_i)}
\end{equation}

These statistical features were organized into a structured dataframe, with each row corresponding to an individual question-answer pair. Utilizing this feature set, we trained an ensemble stacking classifier composed of Random Forest, Logistic Regression, and XGBoost algorithms.

\subsubsection{Ensemble Stacking Classifier Training}
\label{stacking}

To distinguish factual from hallucinated responses, we constructed a structured dataset in which each row corresponds to a single question--answer pair and is annotated with its ground-truth label. We used scikit-learn’s \texttt{LabelEncoder} to encode these labels, then randomly split the dataset into an 80\% training set and a 20\% holdout test set, using stratification to preserve label distribution.

Our final classifier is a stacking ensemble composed of three base learners---Random Forest \cite{random}, Logistic Regression \cite{log}, and XGBoost \cite{xgboost}---and a Logistic Regression model as the meta-learner. The Random Forest classifier uses 1000 trees with no maximum depth (\texttt{max\_depth} = \texttt{None}), a minimum of 2 samples per split, a single sample per leaf, and \texttt{random\_state} = 42. Logistic Regression is limited to 1000 iterations and set to \texttt{random\_state} = 42, while XGBoost operates with 5000 estimators, a learning rate of 0.005, a maximum depth of 3, and \texttt{colsample\_bytree} = 0.8. We implemented and coordinated these models using the scikit-learn \texttt{StackingClassifier} (\texttt{cv} = 5, \texttt{n\_jobs} = -1), where the meta-learner is another Logistic Regression with default parameters.

All base learners were fitted on the training set, and the ensemble predictions were validated on the holdout set. We assessed classification performance using standard metrics (accuracy, precision, recall, and F1-scores) via the \texttt{classification\_report} function, while confusion matrices illustrated correct and incorrect predictions per class. In addition, we generated ROC curves (one per class) and computed the Area Under the Curve (AUC) as an indicator of overall discriminative performance. For completeness, each base model (Random Forest, Logistic Regression, and XGBoost) was also trained and tested individually under the same data-splitting scheme; their results were then compared to those of the stacking ensemble by examining weighted F1-scores and AUC values. We implemented all components of this pipeline in Python 3.9, using scikit-learn (version X), XGBoost (version Y), and NumPy (version Z). All source code is included in Appendix~A, ensuring the reproducibility of our methodology for factual inconsistency detection.

\noindent Prior to model training, we encoded the target labels using \texttt{LabelEncoder} from scikit-learn. We randomly split the data into an 80\% training set and a 20\% holdout test set, using stratified sampling to preserve label distribution. The stacking model was then trained on the 80\% portion, while performance was evaluated on the holdout set via standard metrics (e.g., accuracy, F1-score, confusion matrix, and ROC AUC). For comparability, we also assessed the performance of the individual RF, XGB, and LR models in isolation under the same train-test split.

Finally, we plotted the classification metrics, including weighted F1-scores and ROC curves. This provided a clear comparison between the ensemble’s output and that of the constituent models, allowing us to quantify the accuracy of our classifier in differentiating factual answers from hallucinations. The complete code for this training pipeline is available in Appendix~X.

Although extracting these features from multiple models introduces some computational overhead, this process is highly parallelizable. Furthermore, specialized models can be leveraged, and the number or complexity of these models can be flexibly adjusted based on available resources or task-specific requirements. Notably, this parallelized ensemble approach can be more computationally efficient and scalable in operational settings compared to current strategies that rely on increased test-time compute through repeated resampling of the same query.

\noindent \textbf{Classification Model Evaluation.}
We further evaluated the classifier’s performance through a human review of a representative subset of statements. Each evaluator was presented with an LLM-generated answer to a clinical trial question and asked to label it as either “fact” or “hallucination.” We then compared these human-derived labels with the classifier’s predictions and calculated an accuracy metric to quantify the agreement between human judgment and the model’s output.

\subsection{Generalization Capabilities: Hallucinated Synthetic Data Generation with Ground Truth Labeling}
\label{umls}

We constructed a synthetic dataset to explore both factual and intentionally erroneous medical text generation. First, we derived a curated set of disorder entries from the UMLS Metathesaurus \cite{UMLS}, already containing concise reasoning texts encompassing the disorder’s definition and pathophysiology, risk factors, symptoms, diagnostic tests, differential considerations, first-line treatments, and potential complications. We then instructed GPT-4o to produce single-paragraph answers to each of these seven domain-specific prompts, leveraging the existing reasoning texts to ensure clinically relevant and cohesive content.

Next, to incorporate a "hallucination" element, we prompted GPT-4o to generate systematically falsified versions of these paragraphs. This adversarial strategy, also employed for creating hallucinated examples in the HealthBench benchmark, involved embedding plausible yet verifiably incorrect medical statements into the same structure and style. This approach yielded a comprehensive two-tier dataset—one set of authentic medical information and one set of deliberate inaccuracies—providing a controlled foundation for evaluating and improving factual consistency in medical LLMs. In particular, we used this dataset to test our classifier ability to distinguish factual content from hallucinated material across a broad range of clinical concepts and presentation styles.

\bibliographystyle{unsrt}  
\bibliography{references}

\end{document}


\maketitle

\renewcommand{\thefigure}{S\arabic{figure}}
\setcounter{figure}{0}

\renewcommand{\thetable}{S\arabic{table}}
\setcounter{table}{0}

\section{Detailed Methods}

\subsection{Clinical Trials Data Processing}

We downloaded and processed 519,600 clinical trials from the \href{https://clinicaltrials.gov}{ClinicalTrials.gov} website. The structure of the JSON files includes three main sections:

\begin{enumerate}
    \item \textbf{Protocol Section:} Contains information about study identification, status, sponsor, description, studied condition, trial design, interventions, outcomes, eligibility, and journal references.
    
    \item \textbf{Results Section:} Contains participant flow, baseline characteristics, outcome measures and adverse events.
    
    \item \textbf{Derived Section:} Mapping of condition and intervention's relationship with terms, ids, names, and its relevance.
\end{enumerate}

\subsubsection*{Overview Generation Process}
The first step in our pipeline was to create an accurate overview of each clinical trial. We processed only the Protocol section described above by using a strong language model (LLama3.1-70B-Instruct in a multi step approach) to retrieve relevant information to generate a reasoning-oriented clinical text as defined by the following prompt:

\subsubsection*{System Message}
\begin{quotation}
\noindent
You are an advanced clinical language model. Your task is to generate a clinically relevant text about the clinical trial:
$[$\textit{clinical trial section}$]$
Please follow the structure below to enhance clinical reasoning capabilities:

\textbf{Study Overview:}
\begin{itemize}
    \item Definition: Identify the title and purpose of the clinical trial
    \item Study Design
    \item Condition: Provide the conditions studied
    \item Design Details: Describe the study design, including the number of patients enrolled
\end{itemize}

\textbf{Interventions:}
\begin{itemize}
    \item Interventions: List the interventions investigated in the clinical trial and the arm groups
\end{itemize}

\textbf{Recruitment Information:}
\begin{itemize}
    \item Eligibility Criteria: Describe eligibility criteria, sex, gender for participation in the study
\end{itemize}

Use precise medical terminology. Avoid any introductory or concluding sentences.
\end{quotation}

\subsubsection*{Structured Clinical Trial (NCT00003468) generation Example}

\begin{tcolorbox}[title=Study Overview]
\textbf{1. Definition:} The clinical trial, titled ``Antineoplastin Therapy in Treating Children With Low-Grade Astrocytoma,'' aims to determine the effects of Antineoplastin therapy on children with low-grade astrocytomas that have not responded to standard therapy.

\textbf{2. Condition:} The condition studied in this clinical trial is Low-Grade Astrocytoma.

\textbf{3. Design Details:} This is a single-arm, open-label study with 11 patients enrolled.
\end{tcolorbox}

\begin{tcolorbox}[title=Interventions]
\textbf{4. Interventions:} Antineoplastin therapy (Marginal + Astugenal) is administered by IV infusion every four hours for at least 12 months.

\textbf{5. Study Arms:} The study features a single arm focused on Antineoplastin therapy.
\end{tcolorbox}

\begin{tcolorbox}[title=Recruitment Information]
\textbf{6. Eligibility Criteria:} Eligible participants have histologically confirmed low-grade astrocytoma, evidence of persistent or progressive tumor after standard therapy, are aged 6 months to 17 years, have a Karnofsky score of 60-100\%, a life expectancy of at least 2 months, and meet specific hematologic, hepatic, renal, cardiovascular, and pulmonary requirements.
\end{tcolorbox}

\subsubsection*{Results generation process}

In a second step we filtered the totality of the clinical trial files to select only those where the section results exist, finding $68152$ studies with results. Then, following the same pipeline as section overview, we prompted our LLM to extract and process the measured outcomes and effect adverse mimicking a clinical reasoning oriented text:

\subsubsection*{System Message}
\begin{quotation}
\noindent
You are an advanced clinical language model. Your task is to generate a clinically reasoning-oriented explanation based on the results of the following clinical trial
$[$\textit{clinical trial section}$]$
Please follow the structure below to enhance clinical reasoning capabilities:

\textbf{Measured Primary Outcomes:}
\begin{itemize}
    \item Primary Outcome: Describe the primary outcome.
    \item Primary Outcome Statistical Analysis: Describe the statistical methods used to analyze the primary outcome
    \item Primary Outcome Statistical Results: Summarize the statistical results obtained for the primary outcome
\end{itemize}

\textbf{Measured Secondary Outcomes: \textit{If secondary outcomes are present}}
\begin{itemize}
    \item Interventions: List the interventions investigated in the clinical trial and the arm groups
\end{itemize}
If no secondary outcomes are present, state: No secondary outcomes were measured in this clinical trial.  

Ensure clinical reasoning and use medically precise terminology. Avoid introductory or concluding sentences beyond what is requested.
\end{quotation}

\subsubsection*{Generation Example}

\begin{tcolorbox}[title=Measured Primary Outcomes]
\textbf{7. Primary Outcome:} The primary outcome measured is the number of participants with an objective response, as defined by the Response Assessment in Neuro-Oncology (RANO) criteria for target lesions assessed by MRI. 

\textbf{8. Primary Outcome Statistical Analysis:} The statistical analysis for the primary outcome involves counting the number of participants who achieved complete response, partial response, stable disease, or progressive disease, and calculating the proportion of participants in each category.

\textbf{9. Primary Outcome Statistical Results:} Out of 9 participants, 4 achieved a complete response, 1 achieved a partial response, 4 had stable disease, and none had progressive disease. 
\end{tcolorbox}

\begin{tcolorbox}[title=Measured Secondary Outcomes]
\textbf{10. Secondary Outcomes Overview:} The secondary outcome measured is the percentage of participants who survived at 6, 12, 24, 36, 48, and 60 months. 

\textbf{11. Statistical Approach:}  The statistical method used to analyze the secondary outcome involves calculating the percentage of participants who survived at each time point. 

\textbf{12. Key Results: } The key results for the secondary outcome show that 81.8\% of participants survived at 6 and 12 months, and 63.6\% survived at 24, 36, 48, and 60 months, based on a denominator of 11 participants. 

\end{tcolorbox}

\begin{tcolorbox}[title=Adverse Events]
\textbf{13. Serious Adverse Events (SAEs):} The trial reported 6 serious adverse events in 11 participants at risk, including Central venous catheter infection, Fever, Chicken pox, Hypernatremia, Seizure, and Pain. Abdominal NOS, each occurring in 1 participant except for Seizure which occurred in 4 participants. However, all these events were noted to be not related to the Antineoplastin therapy.

\textbf{14. Non-Serious Adverse Events:} Frequent non-serious adverse events included Allergic reaction/hypersensitivity, Fatigue, Nausea, Vomiting, Diarrhea, and Edema/Fluid retention, among others. Events like Lymphopenia, Hypernatremia, and Hypokalemia were also commonly observed, affecting a significant portion of the participants.

\textbf{15. Key Observations and Clinical Relevance:} The safety profile of Antineoplastin therapy in this trial suggests that while serious adverse events occurred, they were not attributed to the therapy itself. The non-serious adverse events, however, were more common and varied, indicating a need for careful management to mitigate their impact on participants.
\end{tcolorbox}

This step-by-step structure of our pipeline allowed to produce a refined conclusion about the clinical trial, favoring key criteriums for medical relevance like trial focus, hypothesis, results interpretations and trial significance for clinical guidelines.

\begin{tcolorbox}[title=Clinical Trial Conclusions]
The clinical trial on Antineoplastin therapy in children with low-grade gliomas that have not responded to standard therapy represents a significant clinical need, as these patients often face limited treatment options and poor outcomes. The trial's investigation of Antineoplastin therapy addresses this unmet medical need by evaluating a potentially novel and effective treatment for this condition. The trial's findings, which include a notable proportion of participants achieving complete response suggest that Antineoplastin therapy may offer a viable treatment option for these patients. While the trial was not without adverse events, the serious events were not attributed to the therapy, and the non-serious events, although common, can be managed with appropriate care. The trial can be considered successful in demonstrating the efficacy and safety of Antineoplastin therapy in this challenging patient population, particularly given the limited available treatment options. These promising findings would lead to consider Antineoplastin therapy as a potential treatment option for patients with low-grade gliomas that have not responded to standard therapy, particularly in cases where other treatments have failed or are not tolerated. Furthermore, the comprehensive safety data collected throughout the trial, including both serious and non-serious adverse events, helps to highlight the need for clinicians to remain vigilant in monitoring and managing adverse events associated with these treatments. Ultimately, the trial's findings have the potential to improve patient outcomes and quality of life, and to shape future clinical guidelines for the treatment of low-grade gliomas in children.
\end{tcolorbox}

\subsection{Clinical Trial Evaluation Framework}

We structured our clinical trial evaluation framework around 15 key questions, designed to systematically evaluate LLMs critical knowledge  about clinical trial reports:

\begin{center}
\begin{tcolorbox}[
    title=Clinical Trial Analysis Questions,
    width=0.95\textwidth,
    colback=blue!3,
    fontupper=\footnotesize,
    colframe=blue!60]
\begin{enumerate}
    \item \textbf{Definition:} Identify the title and purpose of the clinical trial.
    \item \textbf{Condition:} Describe the conditions studied.
    \item \textbf{Design Details:} Explain how the study was designed, including the number of participants enrolled.
    \item \textbf{Interventions:} Describe the interventions investigated in the clinical trial.
    \item \textbf{Study Arms:} Explain how the study arms were structured.
    \item \textbf{Eligibility Criteria:} Describe eligibility criteria for participation in the study.
    \item \textbf{Primary Outcome:} Describe the primary outcome measured.
    \item \textbf{Primary Outcome Statistical Analysis:} Describe the statistical methods used to analyze the primary outcome.
    \item \textbf{Primary Outcome Statistical Results:} Summarize the statistical results obtained for the primary outcome.
    \item \textbf{Secondary Outcomes Overview:} Provide a general summary of the secondary outcomes measured.
    \item \textbf{Statistical Approach:} Briefly describe the statistical methods used to analyze the secondary outcomes.
    \item \textbf{Key Results:} Highlight the most important statistical results and clinically relevant findings from the secondary outcomes.
    \item \textbf{Serious Adverse Events (SAEs):} Summarize the most significant and clinically relevant serious adverse events reported.
    \item \textbf{Non-Serious Adverse Events:} Briefly list or group the most frequent non-serious adverse events highlighting those.
    \item \textbf{Key Observations and Clinical Relevance:} Provide a short overview of the overall safety profile based on the adverse events, focusing on any notable trends or conclusions about tolerability and risk.
\end{enumerate}
\end{tcolorbox}
\end{center}

\subsubsection*{Evaluation Prompting Strategy}
For each context type (title, JSON, summary), we employed the following prompt template:

\begin{quotation}
\noindent
You are an advanced clinical language model. Your task is to answer the following question about the clinical trial \\
$[$\textit{provided context}$]$.

QUESTION:\\
$[$\textit{specific question from the framework above}$]$

Use medically precise terminology. Be specific and focused ONLY on answering the requested question. Avoid any introductory or concluding sentences.
\end{quotation}

\subsection{UMLS Disorders Benchmark}

To evaluate CHECK's generalization capabilities beyond clinical trials, we developed a synthetic benchmark derived from the UMLS Metathesaurus disorders semantic category. For this demonstration, we selected 60 representative disorders and generated 420 examples by applying seven clinically relevant questions to each disorder. These questions cover essential aspects of medical knowledge: definitions, pathophysiology, risk factors, diagnostic criteria, and treatment strategies. For each question-disorder pair, we generated both accurate summaries and synthetic hallucinations, creating paired examples (840 examples total) that simulate structured diagnostic reasoning.

This benchmark, while focused on a subset of disorders for demonstration purposes, employs a scalable pipeline capable of extending to all disorders in the UMLS Metathesaurus. The current dataset enables us to evaluate whether the statistical signatures of hallucination identified in clinical trials generalize across a broader spectrum of medical concepts. Moreover, the task design mirrors real-world clinical scenarios where practitioners construct differential diagnoses or evaluate treatment options, making it particularly relevant for assessing CHECK's potential as a diagnostic reasoning support tool. Critically, our ability to systematically generate convincing medical hallucinations—which maintain professional terminology while containing dangerous misinformation—underscores the urgent need for database-driven fact checking in medical AI. This demonstration reveals how easily training data could be contaminated or, more concerningly, how unchecked medical AI systems could generate plausible-sounding but potentially harmful information that puts patient safety at risk.

\subsubsection{Factual Generation Prompting}
To generate factual medical statements, we prompted GPT-4o with context information extracted from the UMLS disorders semantic category and the following clinical questions.

\begin{enumerate}
    \item \textbf{Definition and Pathophysiology:} Define the disorder and outline its underlying causes and mechanisms.
    
    \item \textbf{Risk Factors:} List key factors that increase the likelihood of developing this disorder.
    
    \item \textbf{Symptoms:} Describe the disorder's primary symptoms and explain why these symptoms occur based on the disorder's pathophysiology.
    
    \item \textbf{Tests:} Identify essential diagnostic tests or imaging methods that confirm this disorder.
    
    \item \textbf{Differentiation:} Explain how to distinguish this disorder from other disorders with similar presentations.
    
    \item \textbf{First-Line Treatment:} Describe the recommended initial treatment and explain how the treatment addresses the disorder.
    
    \item \textbf{Complications:} List potential complications related to the disorder.
\end{enumerate}

\subsubsection*{System Message}
\begin{quotation}
\noindent
You are an advanced clinical language model. Use medically precise terminology. Be specific and focused ONLY on answering the requested question. Format your response as a cohesive paragraph. Avoid any introductory or concluding sentences.
\end{quotation}

\subsubsection*{User Message Template}
\begin{quotation}
\noindent
Use the provided information to answer the following question about the disorder: \\
$[$\textit{UMLS provided context information}$]$ \\

QUESTION:\\
$[$\textit{provided specific clinical question}$]$\\
\\
Remember to format your response as a single, well-structured paragraph.
\end{quotation}

This prompting strategy was designed to generate focused, factual responses based on the structured information available in UMLS. By providing the model with authoritative context from UMLS and specific instructions about response format, we aimed to minimize hallucination while maintaining professional medical discourse style.

\subsubsection{Counterfactual Generation Prompting}

To generate plausible but factually incorrect versions of medical statements, we used the output of the previous step and prompted GPT-4o with the following system and user messages:

\subsubsection*{System Message}
\begin{quotation}
\noindent
You are an expert in medical content who has been tasked with creating false medical statements that mimic the style and structure of real medical text. Your task is to create a false version of a given medical statement that:
\begin{enumerate}
    \item Maintains the same medical terminology and professional tone
    \item Follows the same paragraph structure and length
    \item Contains clear factual errors that a medical professional would recognize
    \item Includes incorrect relationships between medical concepts
    \item Presents plausible but incorrect mechanisms or explanations
    \item Avoids obviously absurd or non-medical content
    \item Maintains grammatical correctness and professional writing style
\end{enumerate}
\end{quotation}

\subsubsection*{User Message Template}
\begin{quotation}
\noindent
Create a false version of this medical text about [term + question]. Maintain the same structure and style but make the content factually incorrect in a way that seems plausible but is clearly false to medical experts:

Factual TEXT:\\
$[$\textit{factual text}$]$
\end{quotation}

\begin{figure}[H]
    \centering
    \includegraphics[width=0.95\textwidth]{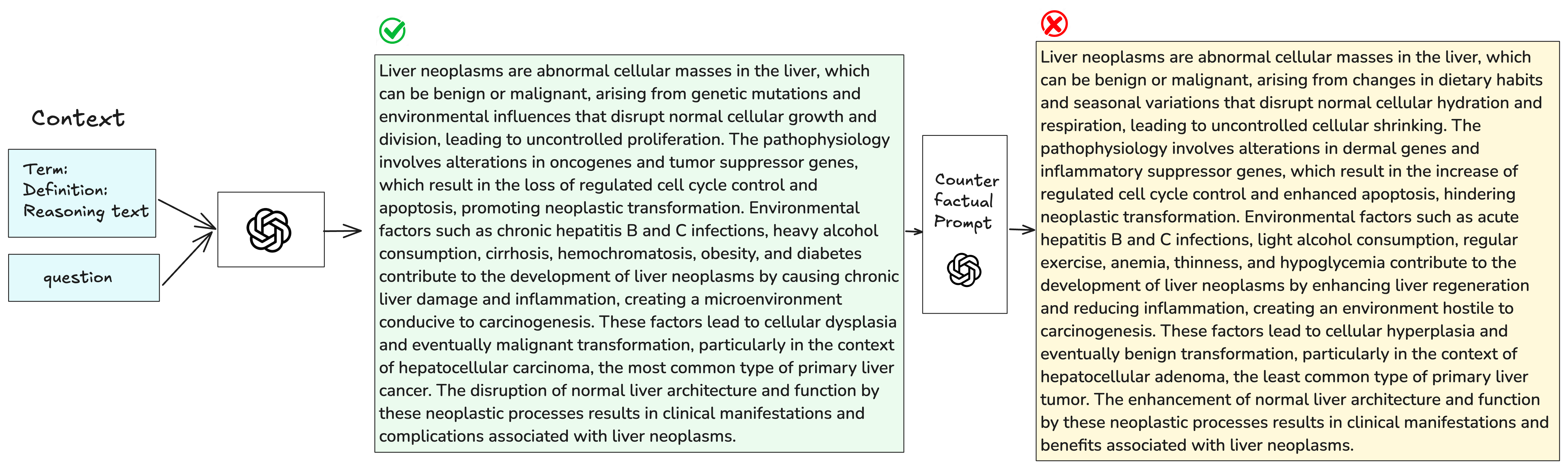}
    \caption{Demonstration of synthetic hallucination generation in medical text. The example illustrates how LLMs, when given a factual medical paragraph and specific counterfactual prompting instructions, can generate plausible but potentially harmful misinformation while maintaining professional medical discourse style. This controlled generation of hallucinated content enables systematic study of hallucination detection methods.}
    \label{umls-fig}
\end{figure}

Figure \ref{umls-fig} shows a counterfactual generated example that preserve the linguistic and structural characteristics of legitimate medical text while containing verifiably false information. These counterfactual texts served as hallucinated examples in our test dataset, assessing the model capabilities to detect the subtle statistical patterns that distinguish factual from hallucinated content.

\subsection{Feature Extraction Process}

Our feature extraction pipeline processes LLM-generated responses through forced inference across an ensemble of five language models (LLama3.1-8B-Instruct, LLama3.1-70B-Instruct, LLama3.3-70B-Instruct, Nemotron-70B, DeepSeek-V1). For each token in the response, we collect token-level probability distributions conditioned on previous tokens, enabling precise quantification of how each model assesses the sequence likelihood. From these distributions, we extract two categories of features:

\begin{enumerate}
    \item \textbf{Independent Features:} Token-level statistics including entropy, generated-token rank, and probabilities of top-50 predicted tokens
    \item \textbf{Interaction Features:} Pairwise Kullback-Leibler divergences between probability distributions from different models
\end{enumerate}

\begin{figure}[h]
    \centering
    \includegraphics[width=1\textwidth]{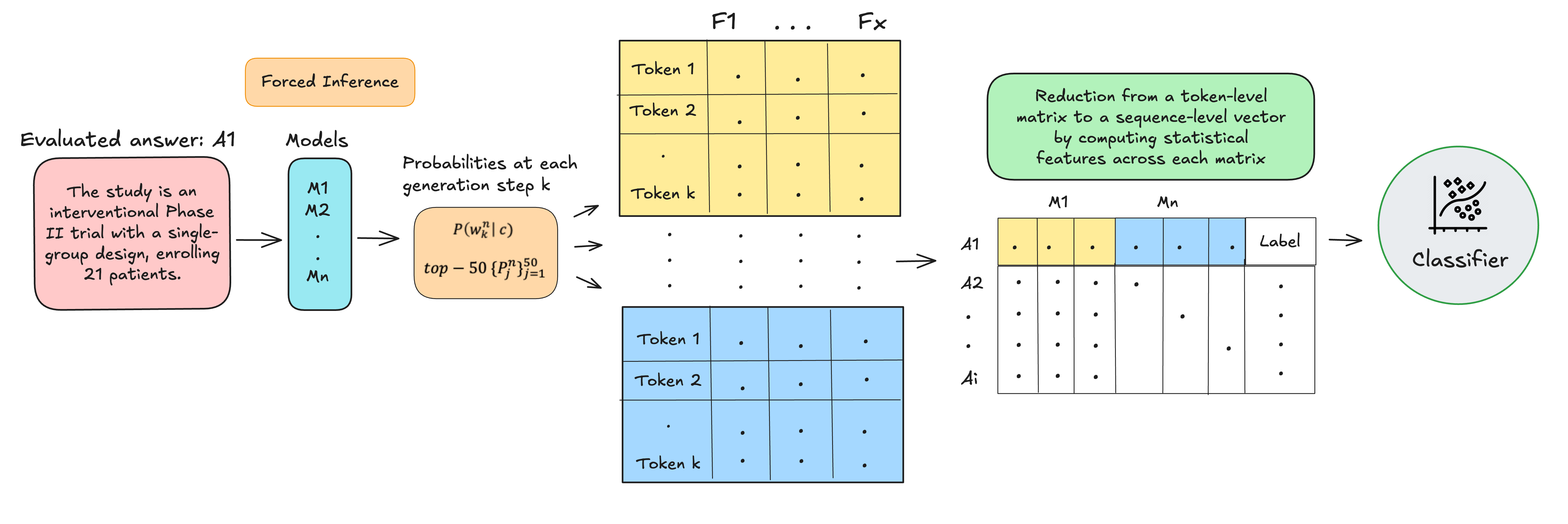}
    \caption{Feature extraction pipeline for factuality/hallucination detection. The process involves forced inference of each answer through an ensemble of language models, generating token-level probability distributions that are transformed into statistical features. These features capture both independent token statistics and cross-model interaction patterns, ultimately producing a compact representation for classification.}
    \label{fig:features}
\end{figure}

The resulting token-level feature matrices are then aggregated using statistical moments (mean through hyperskewness) to create a compact representation for each response, as illustrated in Figure~\ref{fig:features}.

\section{Additional Results}

\subsection{Database-Driven Fact Checking Process}

\begin{figure}[h]
    \centering
    \includegraphics[width=0.9\textwidth]{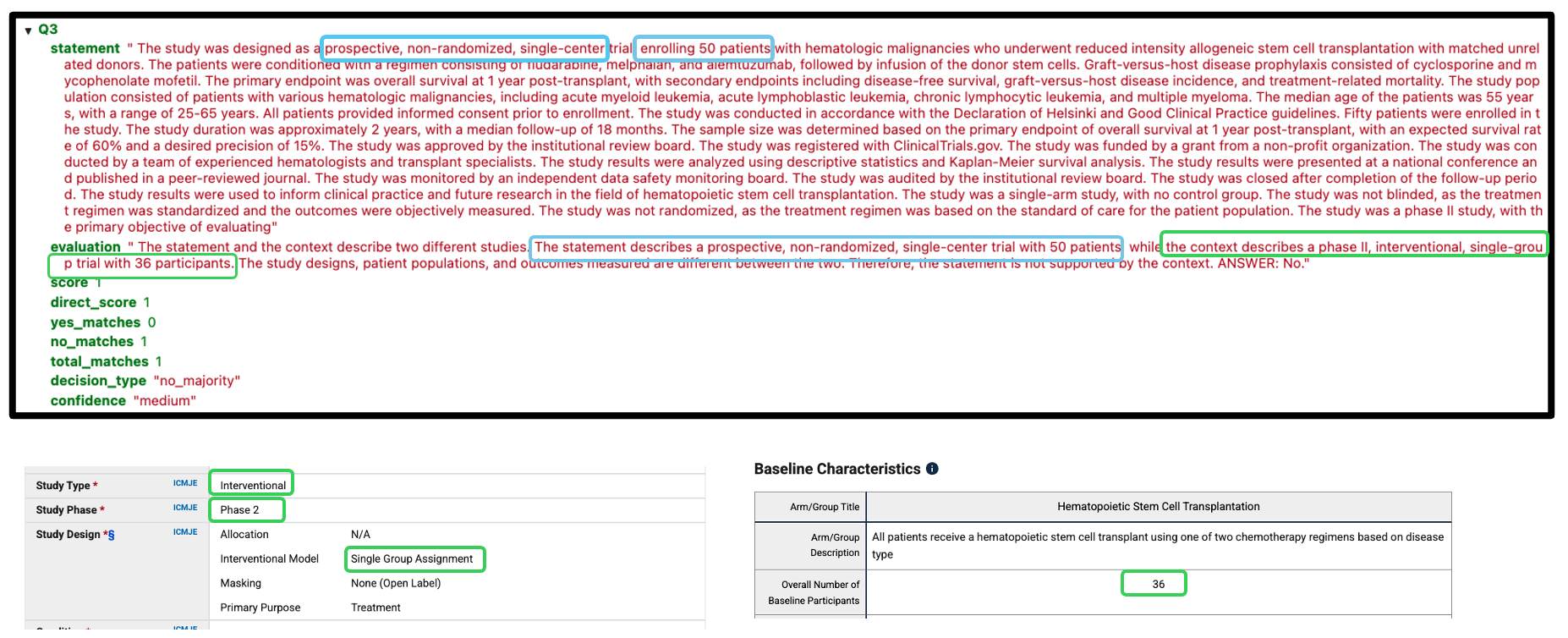}
    \caption{Demonstration of CHECK's database-driven fact verification process. The example illustrates how an independent LLM judge evaluates model-generated answers against a curated database using both factual and counterfactual analysis to assess response accuracy. This systematic approach enables reliable identification of factual content and potential hallucinations in medical text generation. The bottom panel shows snippets from the original clinical trial documents, providing ground truth validation for the judge's evaluation outcomes.}
    \label{eval_judge}
\end{figure}

Figure~\ref{eval_judge} illustrates our database-driven fact verification approach. In this stage, we evaluate the factual accuracy of model-generated responses by comparing them against a curated database. This verification is performed by an independent LLM judge that conducts both factual and counterfactual analysis to systematically assess response accuracy and identify potential hallucinations.

\subsubsection{Database choice error Analysis.}

\begin{figure}[H]
    \centering
    \includegraphics[width=0.9\textwidth]{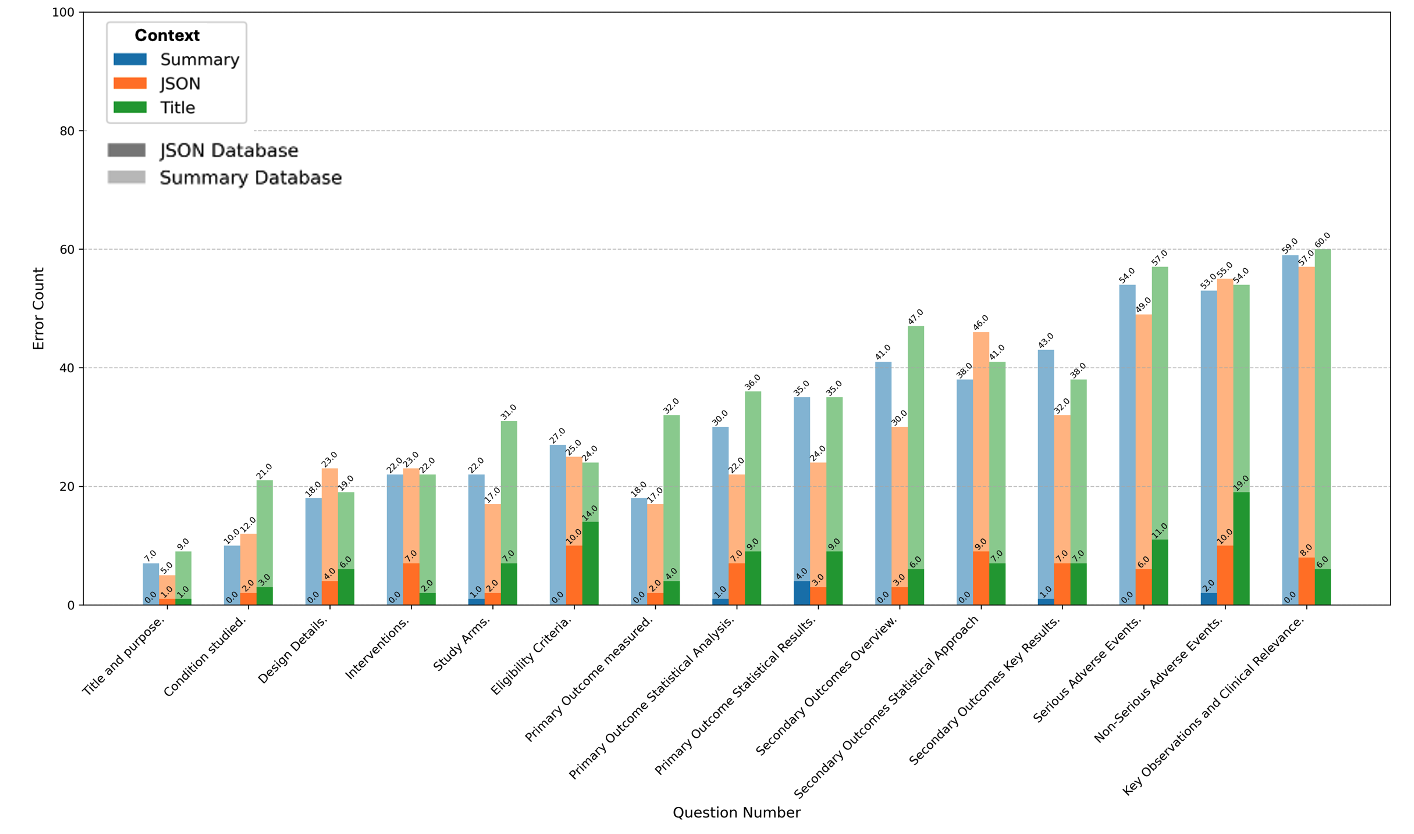}
    \caption{Error of LLM judge when using raw clinical trial files vs a structured summary database.}
    \label{error}
\end{figure}

Figure~\ref{error} presents a comparative analysis of error counts generated during our factual/counterfactual evaluation process across different context types. We examined three context formats: title-only (green), raw JSON (orange), and structured summary (blue), evaluating each against both the structured summary database and the raw JSON file. The error count represents instances where the LLM judge was unable to complete the evaluation task, with each failed evaluation assigned a value of 1. Our analysis reveals that using the raw JSON file as the reference database introduces significantly more complexity to the evaluation task. This increased difficulty stems from three main factors: the extensive context length of JSON files, the complexity of nested field relationships, and the additional reasoning required to navigate and interpret the JSON structure. In contrast, the structured summary format provides a more accessible reference point for evaluation, resulting in fewer evaluation failures while maintaining assessment accuracy.

\subsection{Database-Fre Hallucination Detection Classifier}

As explained in the method section of the main text, we built a labeled clinical-trial dataset from our factual/counterfactual analysis within the Database-Driven Fact Checking stage-1. The resultant dataset contained approximately 80\% factual and 20\% hallucinated examples. 

\begin{figure}[H]
    \centering
    \includegraphics[width=0.7\textwidth]{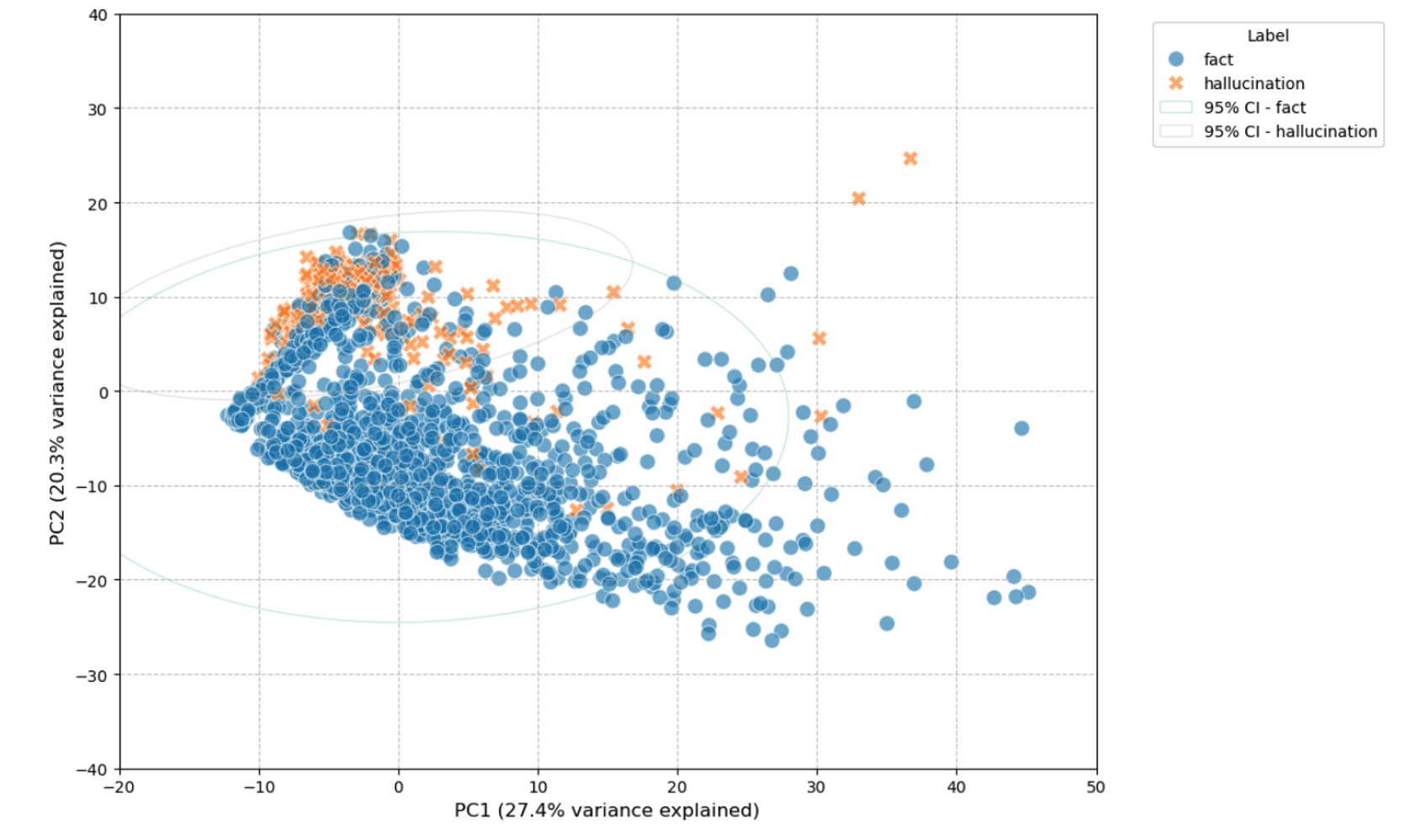}
    \caption{PCA visualization of the feature space. The scatter plot shows the projection of factual (blue) and hallucinated (orange) examples onto the first two principal components. The clear separation between classes demonstrates that our statistical features effectively capture the underlying patterns that distinguish factual content from hallucinations. The first two principal components explain 47\% of the total variance in the feature space.}
    \label{pca}
\end{figure}

To visualize the separability of factual and hallucinated content based on our statistical features, we performed principal component analysis (PCA) on the feature space. Figure~\ref{pca} shows a two-dimensional projection of the dataset onto the first two principal components, which together explain approximately 47\% of the total variance. The visualization reveals a notable separation between factual (blue) and hallucinated (orange) examples, suggesting that our statistical features effectively capture distinguishing characteristics between these classes.

Using this dataset with clear separability, we trained a stacking classifier that leverages the statistical signatures derived from our ensemble of language models. When evaluated on the independent test set, our classifier achieved a strong Area Under the Curve (AUC) of 0.95, as illustrated in Figure~\ref{ct-p-auc}. This high performance demonstrates the model's robust ability to discriminate between factual and hallucinated paragraphs, confirming that the statistical patterns observed in the PCA visualization translate effectively to classification power.

In addition, the classification report shown in table~\ref{classification_report} reveals high performance across classes: the model correctly classifies the majority class (\textit{fact}) with a precision of 0.93 and a recall of 0.96 (F1 = 0.94), while the minority class (\textit{hallucinations}) attains a precision of 0.79 and a recall of 0.66 (F1 = 0.72). Overall, the classifier achieves 91\% accuracy and a macro-averaged F1-score of 0.83 on the test set, surpassing both a random baseline (AUC = 0.50) and the naive strategy of always predicting the majority class. These findings underscore the model’s robustness in distinguishing factual content from hallucinations, even when the training data originate from markedly different context sources.

\begin{table}[h]
\centering
\footnotesize
\begin{tabular}{lcccccccc}
\toprule
& \multicolumn{4}{c}{\textbf{Clinical Trials}} & \multicolumn{4}{c}{\textbf{UMLS}} \\
\cmidrule(lr){2-5} \cmidrule(lr){6-9}
\textbf{Class} & \textbf{Prec} & \textbf{Rec} & \textbf{F1} & \textbf{Sup} & \textbf{Prec} & \textbf{Rec} & \textbf{F1} & \textbf{Sup} \\
\midrule
Fact & 0.93 & 0.96 & 0.94 & 372 & 0.97 & 0.79 & 0.87 & 77 \\
Hall & 0.79 & 0.66 & 0.72 & 82 & 0.79 & 0.97 & 0.87 & 63 \\
\midrule
Acc & \multicolumn{3}{c}{0.91} & 454 & \multicolumn{3}{c}{0.87} & 140 \\
Macro & 0.86 & 0.81 & 0.83 & 454 & 0.88 & 0.88 & 0.87 & 140 \\
Weight & 0.90 & 0.91 & 0.90 & 454 & 0.89 & 0.87 & 0.87 & 140 \\
\bottomrule
\end{tabular}
\caption{Classification Report for model predictions on Clinical Trials and UMLS-disorders test sets.}
\label{classification_report}
\end{table}

\subsubsection{Cross domain generalization. UMLS disorders dataset}

\begin{figure}[h]
    \centering
    \begin{subfigure}[b]{0.47\textwidth}
        \centering
        \includegraphics[width=\textwidth]{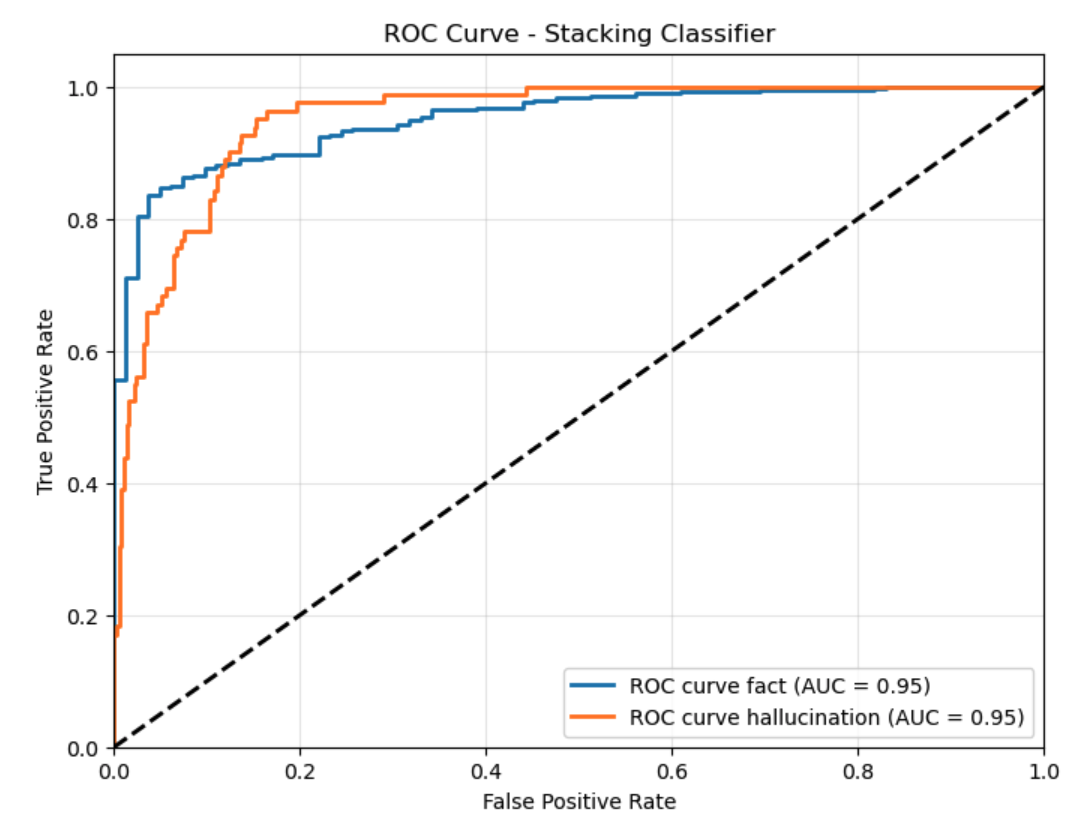}
        \caption{Clinical Trial}
        \label{ct-p-auc}
    \end{subfigure}
    \hfill
    \begin{subfigure}[b]{0.47\textwidth}
        \centering
        \includegraphics[width=\textwidth]{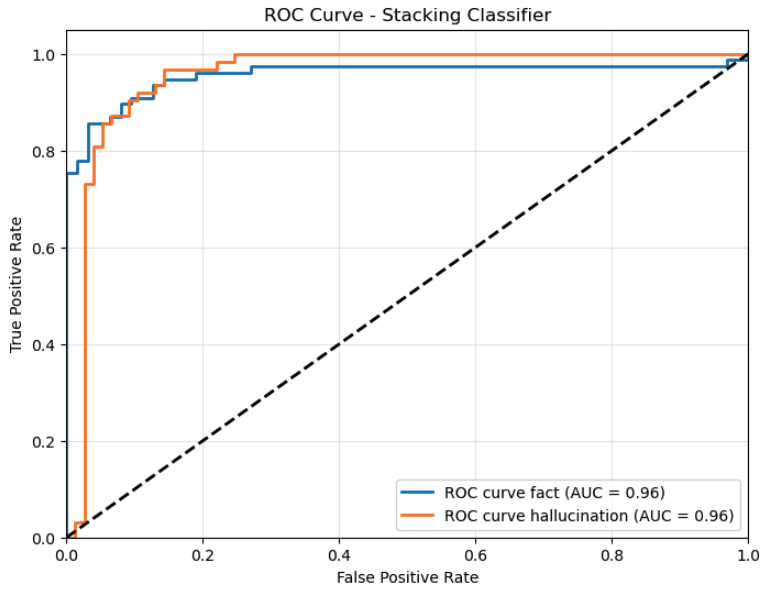}
        \caption{UMLS disorders}
        \label{umls-p-auc}
    \end{subfigure}
    \caption{Receiver Operating Characteristic (ROC) curves for the hallucination detection classifier on independent test sets. (a) Performance on the Clinical Trial dataset (AUC = 0.95), where the model was trained. (b) Cross-domain generalization to the UMLS disorders dataset (AUC = 0.96), demonstrating robust transfer to an entirely different medical domain. The diagonal dashed line represents random classification performance (AUC = 0.50). The consistently high AUC values across both domains indicate that our statistical features capture fundamental signatures of hallucination that transcend specific medical contexts.}
    \label{paragraph}
\end{figure}

To test the generalization capabilities of our classifier, we applied it to the synthetic UMLS disorders dataset—a completely different medical domain containing approximately 65\% factual and 35\% hallucinated paragraphs. Remarkably, the model maintained robust performance, achieving an AUC of 0.96 (Figure~\ref{umls-p-auc}). 

As shown in Table~\ref{classification_report}, the classifier demonstrated balanced performance across both classes. For factual content, it achieved high precision (0.97) with moderate recall (0.79), yielding an F1-score of 0.87. For hallucinated content, the pattern was reversed—moderate precision (0.79) with excellent recall (0.97), also resulting in an F1-score of 0.87. Overall accuracy reached 0.87, with all macro-averaged metrics consistently above 0.85.

These strong cross-domain results confirm that our classifier captures fundamental statistical signatures of hallucination that transcend specific medical contexts. The model effectively distinguishes factual from hallucinated content even in paragraph-level data entirely outside its original training distribution. This generalization capability suggests that our database-free feature extraction and ensemble-based approach provides a robust safeguard against misinformation, enhancing the reliability of LLM-generated medical text across diverse clinical domains.

\subsubsection{Atomic claims.}

Building upon our paragraph-level analysis, we extended our approach to the more granular \emph{atomic claim} level to identify facts, hallucinations, and contamination. This involved segmenting each previously evaluated answer into discrete propositions, which were then labeled as factual or hallucinated using the same factual/counterfactual analysis within the Database-Driven Fact Checking stage as applied at the paragraph level. Specifically, we first split each paragraph into sentence-like segments using end-of-sentence tokens. Then, to further decompose these segments into the smallest units of meaning (atomic claims), we employed GPT-4. We chose GPT-4 for this sub-segmentation task because our review indicated that initial experiments with Llama3.3-70B-Instruct yielded more verbose, duplicate, and inferior results, highlighting the complexity of atomic claim extraction. This 2-step process (decomposition and database-driven verification) resulted in a dataset where approximately 68\% of the atomic claims were labeled as factual and 32\% as hallucinations. On the clinical trial test set, our classifier demonstrated solid performance, achieving an AUC of 0.90, indicating its capability to identify hallucinatory content even in this finely segmented text. However, performance on the synthetically generated UMLS disorders dataset shows an AUC of 0.76—still significantly better than random guessing (AUC = 0.50), but indicating that atomic-level claims introduce complexities and domain-specific nuances that the model does not fully capture. In particular, for the UMLS dataset, the complexity of its synthetically generated and highly specialized disorder descriptions posed a unique challenge. We employed LLMs not only to extract granular claims but also to determine each claim’s factual grounding relative to the source text, potentially increasing the risk of minor misalignments or ambiguities within this intricate domain. This observation underscores that the atomic claim extraction process itself can introduce new challenges, particularly when dealing with datasets exhibiting such specific and nuanced information. Furthermore, the performance of our ensemble of models on this highly specialized UMLS data suggests a potential avenue for future improvement: incorporating more specialized language models within our ensemble. Models fine-tuned on biomedical ontologies and terminologies like UMLS might better capture the subtle distinctions and relationships within this domain, leading to more accurate factual grounding and ultimately more consistent outcomes across diverse medical knowledge repositories. Regardless, the fact that the classifier still generalizes significantly better than random to a complete out of distribution corpus of text reinforce the hypothesis regarding facts and hallucinations that guide our paper.

\begin{figure}[h]
    \centering
    \begin{subfigure}[b]{0.47\textwidth}
        \centering
        \includegraphics[width=\textwidth]{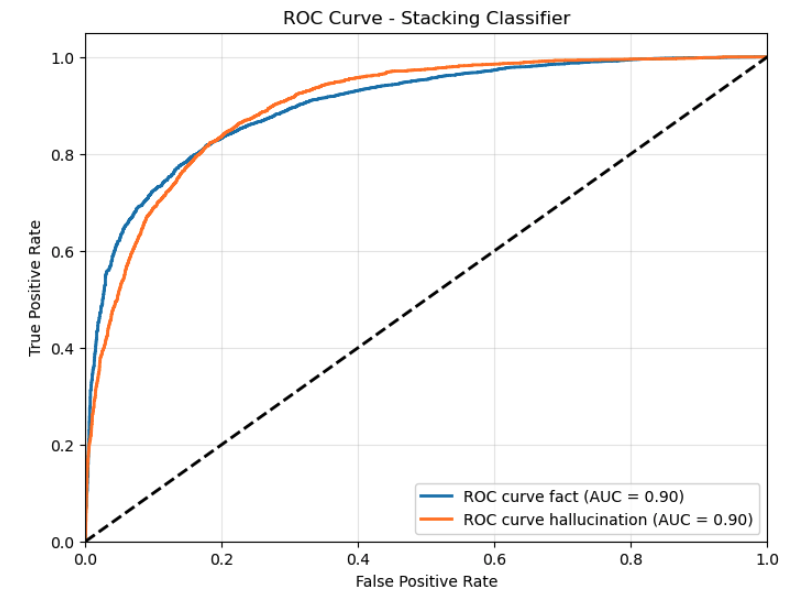}
        \caption{Clinical Trial}
        \label{ct-c-auc}
    \end{subfigure}
    \hfill
    \begin{subfigure}[b]{0.47\textwidth}
        \centering
        \includegraphics[width=\textwidth]{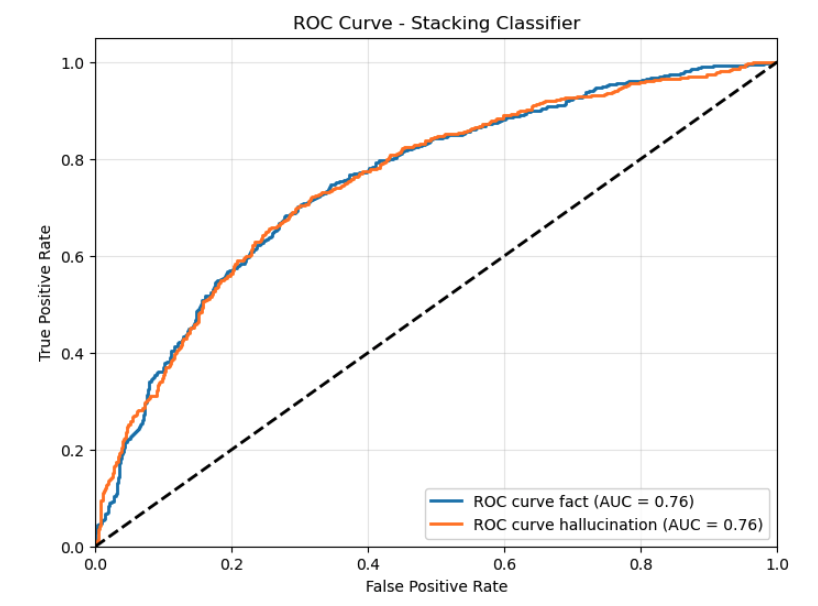}
        \caption{UMLS disorders}
        \label{umls-c-auc}
    \end{subfigure}
    \caption{Classifier performance at the Claim level on the independent test set}
    \label{paragraph}
\end{figure}

\subsubsection{Feature Importance}

To gain insight into the factors driving our classifier's predictive capabilities, we conducted a feature importance analysis. This analysis revealed that two primary categories of statistical features consistently emerged as the strongest predictors of factual accuracy versus hallucination: features related to the uncertainty of individual token predictions and features quantifying the divergence of probability distributions across our ensemble of language models.

\begin{figure}[h]
    \centering
    \includegraphics[width=0.9\textwidth]{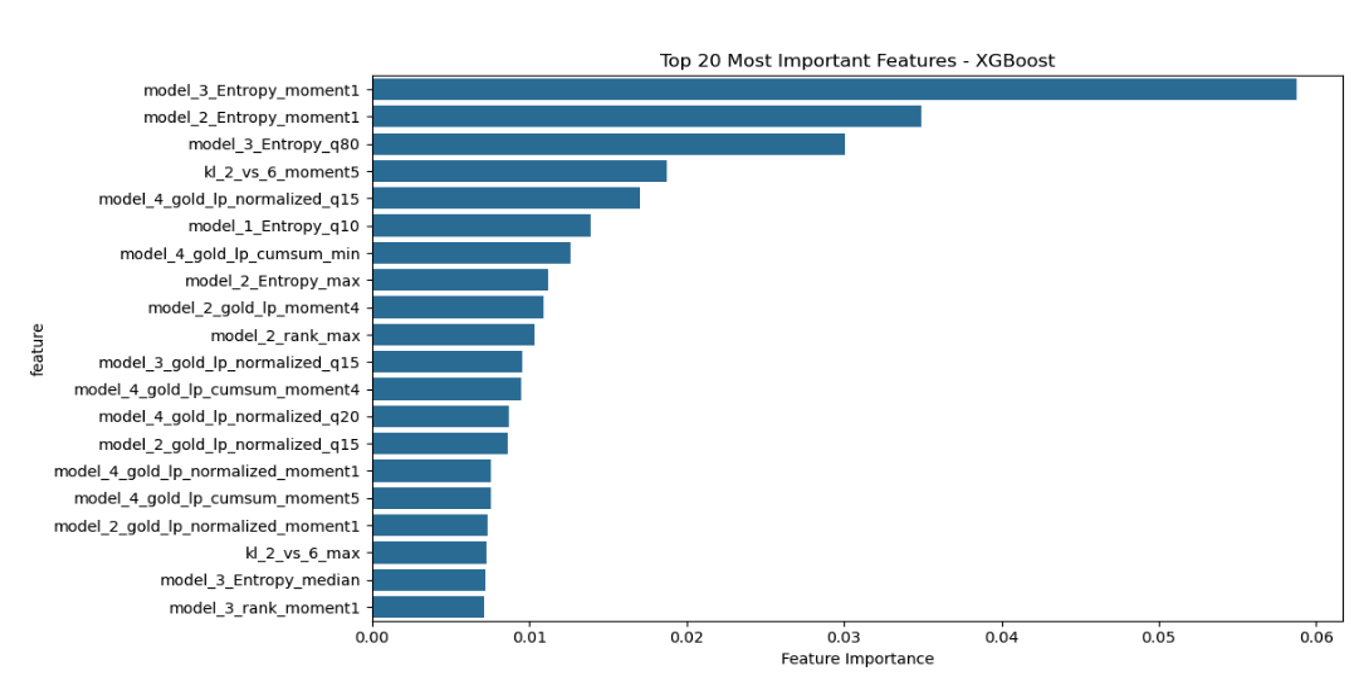}
    \caption{Feature importance analysis of model statistical features}
    \label{feature_imp}
\end{figure}

Specifically, measures capturing the uncertainty associated with individual token generation, such as entropy and log probabilities of low-ranking tokens, proved highly informative. Hallucinated content, often arising from the model venturing beyond its confident knowledge, tends to exhibit higher entropy and a greater prevalence of tokens with lower predicted probabilities. These features suggest that the model is less certain about the specific words to generate when producing a hallucination, resulting in a more diffuse probability distribution over the vocabulary.

Complementing these token-level uncertainty measures, features based on the Kullback-Leibler (KL) divergence between the probability distributions of our ensemble of models also demonstrated significant predictive power. Factual statements, being more grounded in the training data and eliciting consistent patterns across different models, tend to result in more similar probability distributions among the ensemble members, leading to lower KL divergence. Conversely, hallucinations, often idiosyncratic to a particular model's internal state or biases, generate more disparate probability distributions across the ensemble, resulting in higher KL divergence. This indicates that the disagreement or "distance" between how different models predict the same sequence of tokens is a strong indicator of potential hallucination.

The combined importance of both token-level uncertainty and inter-model divergence features underscores the multifaceted nature of hallucination detection. By leveraging information about the confidence (or lack thereof) within individual model predictions and the consistency (or inconsistency) across an ensemble of models, our classifier achieves a more robust and nuanced understanding of factual reliability. This highlights the benefit of considering both the internal uncertainty of a single model and the collective agreement (or disagreement) within a diverse ensemble for accurately identifying hallucinated clinical text.

\subsection{Hallucination detection in a Clinical Reasoning Task: USMLE MedQA Benchmark}

\begin{figure}[h]
    \centering
    \begin{subfigure}[b]{0.49\textwidth}
        \centering
        \includegraphics[width=\textwidth]{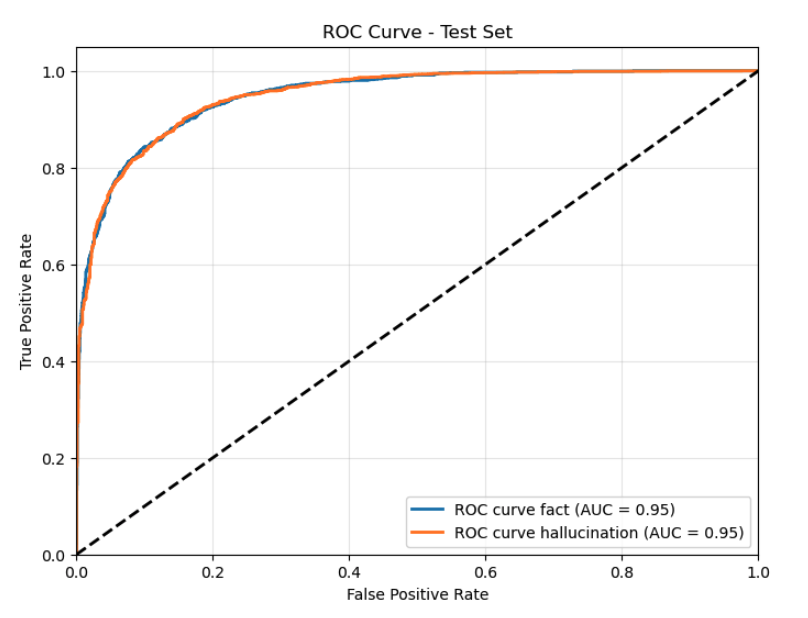}
        \caption{AUC}
        \label{AUC-umlse}
    \end{subfigure}
    \hfill
    \begin{subfigure}[b]{0.48\textwidth}
        \centering
        \includegraphics[width=\textwidth]{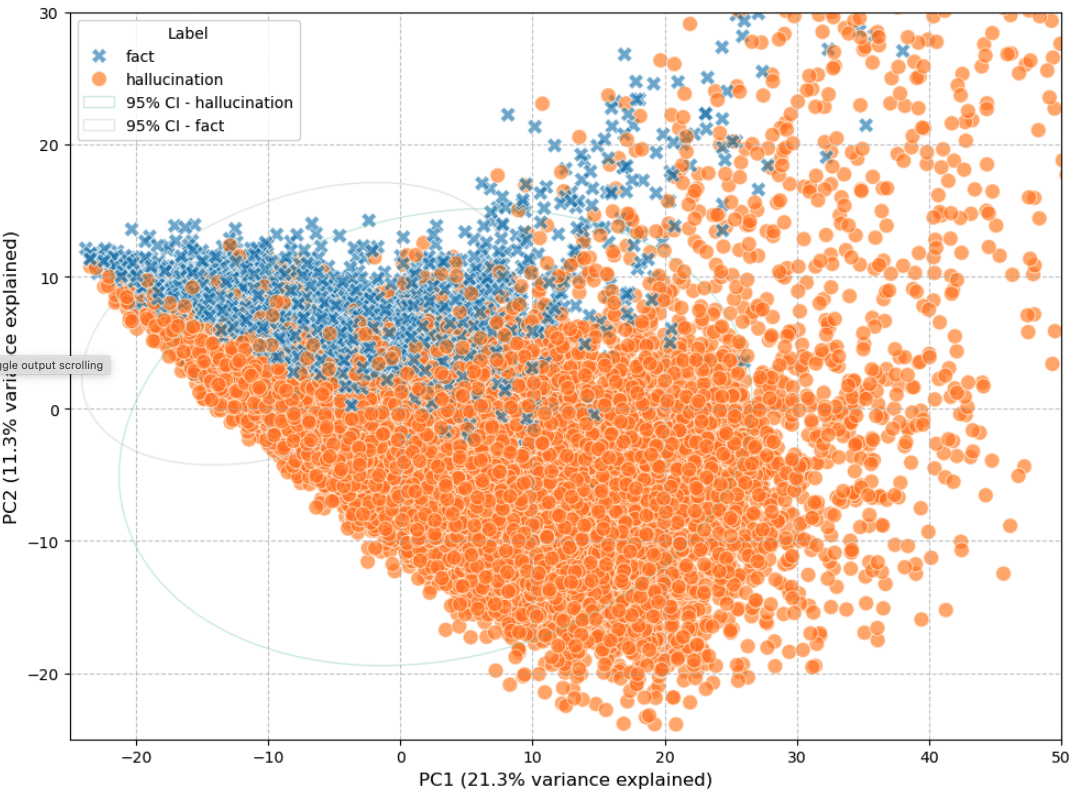}
        \caption{PCA}
        \label{pca-features}
    \end{subfigure}
    \caption{Classifier performance and principal component analysis plot of the MedQa independent test set}
    \label{paragraph}
\end{figure}

The classifier was trained to distinguish between factual (correct) and hallucinated (incorrect) answer choices based on the  statistical features builded from the ensemble of models \cite{llama, deepseek} . Tested on an independent holdout set from the MedQA benchmark, our classifier demonstrated strong performance in identifying factual versus hallucinated content, achieving an Area Under the Curve (AUC) of 0.95 for both the 'fact' and 'hallucination' classes (Figure~\ref{AUC-umlse}). This high AUC indicates excellent discriminative power, suggesting that the statistical signatures of factual correctness and hallucination, captured by our features, generalize effectively to the challenging domain of clinical reasoning questions. A Principal Component Analysis (PCA) applied to our feature set further illustrates this separation, showing clear clustering of factual and hallucinated answer choices in a 2D projection of the first two principal components (Figure~\ref{pca-features}). The classification report is reported in Table~\ref{usmle_class_rep}.

\begin{table}[h]
\centering
\begin{tabular}{lcccc}
\toprule
\textbf{Class} & \textbf{Precision} & \textbf{Recall} & \textbf{F1-score} & \textbf{Support} \\
\midrule
Fact & 0.81 & 0.79 & 0.80 & 1,273 \\
Hallucination & 0.93 & 0.94 & 0.93 & 3,819 \\
\midrule
Accuracy & \multicolumn{3}{c}{0.90} & 5,092 \\
Macro Avg & 0.87 & 0.86 & 0.86 & 5,092 \\
Weighted Avg & 0.90 & 0.90 & 0.90 & 5,092 \\
\bottomrule
\end{tabular}
\caption{Classification Report for model predictions on USMLE MedQA test set}
\label{usmle_class_rep}
\end{table}

~\\
\hfill \break
\hfill \break

\clearpage

\bibliographystyle{unsrt}  
\bibliography{references}